\documentclass[a4paper,fleqn]{cas-dc}

\usepackage[authoryear]{natbib}
\usepackage{hyperref}
\usepackage{color,soul}

\def\tsc#1{\csdef{#1}{\textsc{\lowercase{#1}}\xspace}}
\tsc{WGM}
\tsc{QE}

\begin{document}
\let\WriteBookmarks\relax
\def\floatpagepagefraction{1}
\def\textpagefraction{.001}

\shorttitle{Gap Completion in Point Cloud Scene occluded by Vehicles using SGC-Net}    

\shortauthors{Feng et al.}  

\title [mode = title]{Gap Completion in Point Cloud Scene occluded by Vehicles using SGC-Net}



%

\author[1]{Yu Feng}[orcid=0000-0001-5110-5564]\fnref{fn1}\corref{cor1}
\author[2]{Yiming Xu}\fnref{fn1}
\author[3]{Yan Xia}
\author[2]{Claus Brenner}
\author[2]{Monika Sester}

\cortext[cor1]{Corresponding author}
\fntext[fn1]{These authors contribute equally to this work.}

\affiliation[1]{organization={Chair of Cartography and Visual Analytics, Technical University of Munich},
addressline={Arcisstrasse 21}, 
city={Munich},
postcode={80333}, 
country={Germany}}

\affiliation[2]{organization={Institute of Cartography and Geoinformatics, Leibniz University Hannover},
addressline={Appelstra\ss{}e 9a}, 
city={Hannover},
postcode={30167}, 
country={Germany}}

\affiliation[3]{organization={Computer Vision Group, Technical University of Munich},
addressline={Boltzmannstrasse 3}, 
city={Garching},
postcode={85748}, 
country={Germany}}














\begin{abstract}
Recent advances in mobile mapping systems have greatly enhanced the efficiency and convenience of acquiring urban 3D data.
These systems utilize LiDAR sensors mounted on vehicles to capture vast cityscapes. 
However, a significant challenge arises due to occlusions caused by roadside parked vehicles, leading to the loss of scene information, particularly on the roads, sidewalks, curbs, and the lower sections of buildings.
In this study, we present a novel approach that leverages deep neural networks to learn a model capable of filling gaps in urban scenes that are obscured by vehicle occlusion.
We have developed an innovative technique where we place virtual vehicle models along road boundaries in the gap-free scene and utilize a ray-casting algorithm to create a new scene with occluded gaps.
This allows us to generate diverse and realistic urban point cloud scenes with and without vehicle occlusion, surpassing the limitations of real-world training data collection and annotation. 
Furthermore, we introduce the Scene Gap Completion Network (SGC-Net), an end-to-end model that can generate well-defined shape boundaries and smooth surfaces within occluded gaps. 
The experiment results reveal that 97.66\% of the filled points fall within a range of 5 centimeters relative to the high-density ground truth point cloud scene.
These findings underscore the efficacy of our proposed model in gap completion and reconstructing urban scenes affected by vehicle occlusions.
\end{abstract}




\maketitle

\section{Introduction}
\label{sec:Introduction}


With the rapid development of 3D acquisition technology, a wide range of 3D sensors have emerged, including LiDAR systems mounted on vehicles or aircraft, as well as RGB-D cameras. 
These sensors enable the rapid acquisition of vast quantities of 3D data.
The availability of 3D data provides a wealth of rich geometric and volumetric information, surpassing what can be obtained from single 2D images.

LiDAR mobile mapping systems enable the efficient collection of such data. 
The measured point clouds are often used for different purposes, such as road boundary extraction \citep{zhou2012mapping,yang2013semi}, lane marking extraction \citep{mi2021two}, 3D building refinement~\citep{wysocki2022refinement}, terrain modelling~\citep{feng2018enhancing}.
With these output products, different applications have been realized, including the generation of HD-Maps for autonomous driving~\citep{bao2023review}, planning for solar systems~\citep{biljecki2015propagation}, and also city's emergency response~\citep{feng2022determination}.

\begin{figure}[h]
\centering
\includegraphics[width=0.48\textwidth]{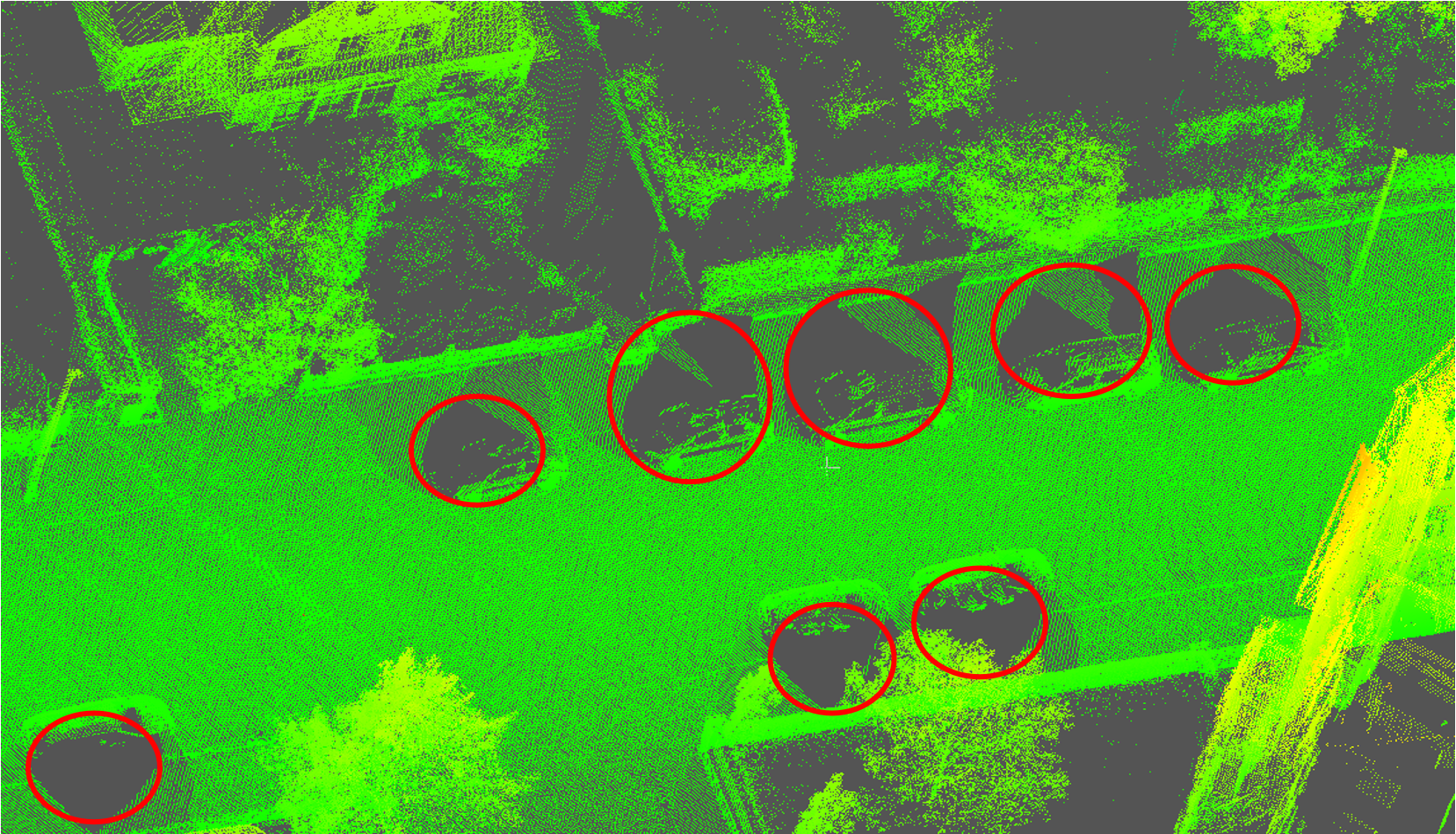}
\caption{Point cloud gaps resulting from vehicles parked alongside the road.}
\label{fig:gap_examples}
\end{figure}

However, the data usually contains extraneous dynamic objects that are inherently not part of the static environment, such as vehicles, bicycles, and pedestrians. 
Such interference may hinder the completeness of point cloud data, often requiring manual or semi-automatic post-processing to compensate for inaccurate modeling caused by occlusion.

Among these objects, vehicles stand out as the most significant contributors in terms of occluded area \citep{barros2022visibility}. 
As illustrated in Figure \ref{fig:gap_examples}, parked vehicles along the roadside can obscure ground surfaces, walls, and adjacent building facades. 
With point cloud data that has many such gaps, researchers may need to develop specific algorithms for individual applications, such as closing the gaps on terrain surfaces using interpolation \citep{feng2018enhancing}, completing the road boundary using curve fitting \citep{wang2015road}.


This work focuses on tackling this challenge by leveraging contextual information surrounding the point cloud gaps and employing an idea similar to the inpainting techniques used in 2D image processing. 
The point cloud gaps primarily occur in close proximity to the lower sections of the roads due to vehicle occlusion.
These gaps encompass various scenarios, such as road boundaries between sidewalks and streets, delineations of different ground surface types, the intricate shapes of grasses and bushes, as well as the geometry of absent vertical walls or building facades near the ground. 
This work offers substantial benefits to downstream tasks, specifically in the modeling of fine-grained terrain surfaces and road boundaries. It also enables seamless visualization of 3D scenarios without the interference of dynamic objects, enhancing the overall accuracy and realism of the visual representations.

This work provides the following two contributions. 
First, we have developed an innovative technique, where we place virtual vehicle models along road boundaries in the gap-free scene and utilize a ray-casting algorithm to create a new scene with gaps caused by occlusion.
This allows us to generate a diverse and realistic urban point cloud scene with and without vehicle occlusion, surpassing the limitations of real-world training data collection for learning tasks. 
Second, we introduce a novel Scene Gap Completion Network (SGC-Net), an end-to-end model to generate well-defined shape boundaries and smooth surfaces for gaps caused by occlusion. 
In the evaluations, we show that our model outperforms several baseline methods.

The subsequent sections of this paper are structured as follows:
Section 2 provides a comprehensive overview of existing works concerning point cloud and shape completion techniques.
In Section 3, we introduce our workflow for data processing of point clouds step-by-step, encompassing data pre-processing and the generation of training datasets through ray-casting.
Section 4 delves into the network architecture design employed in this study.
In Section 5, our method is evaluated on both synthetic datasets and real-world scenarios, providing a comprehensive assessment of its performance and applicability.
Finally, in Section 6, we conclude the paper and provide an outlook.

\section{Related Work}
\label{sec:Related Work}
The completion of 3D shapes is a problem of great interest and has been extensively studied in computer vision and robotics. 
As summarized by \citet{xia2021vpc}, there are in general three types of methods for shape completion, namely geometry-based, template-based, and learning-based methods.

Geometry-based methods complete the gaps in point clouds by leveraging the inherent geometric properties of common objects, such as ensuring surface continuity and smoothness or symmetry of human-made objects. 
While these methods are usually effective for addressing small holes, they encounter challenges when attempting to fill larger gaps.

Template-based methods, such as \cite{nan2010smartboxes, nan2012search} involve filling in the gaps by searching for the most similar template from a provided 3D shape database. They then assign new points based on the retrieved surface information. These methods tend to be time-consuming due to the retrieval process and can be labor-intensive when constructing the database.

As a result, learning-based methods have therefore become increasingly popular in recent years, particularly when dealing with large-scale 3D datasets. 
These methods have emerged as the dominant approach in various fields, leveraging their capabilities to tackle complex problems and deliver remarkable results.
Given that such methods typically encompass two primary components, namely 3D feature extraction and point cloud generation, we summarize the relevant research in the following based on these two aspects.
It is important to note that this is a very active research field, and for
a more comprehensive overview of the related literature, readers are encouraged to refer to \cite{fei2022comprehensive}.

\subsection{3D Feature Encoders}
\label{sec:Feature_Extraction}
The prerequisite of point cloud completion is to be able to extract features from the surface and shape of the point cloud. 
Feature extraction aims to capture meaningful global and local information from point cloud data. 
PointNet \citep{qi2017pointnet} and its variants have emerged as the most popular approach for feature extraction from point clouds. 
These methods directly process unordered sets of points as input, eliminating the need for preprocessing steps, such as voxelization.
The PointNet architecture utilizes shared multi-layer perceptrons (MLPs) and max pooling operations to learn informative features from the point cloud data.

An enhanced method named PointNet++ \citep{qi2017pointnet++} has been developed to capture both local and global features from point clouds through hierarchical feature learning. 
Since point clouds lack order and exhibit irregular local arrangements, using a graph structure becomes ideal for extracting local features and global information from the point cloud.


PointCNN \citep{li2018pointcnn} introduces a novel kernel function that directly applies convolutional neural networks (CNNs) to point clouds. 
This kernel function can be used to define the convolution operation of unordered point sets to obtain point cloud features. 
PointSIFT \citep{jiang2018pointsift} and SOE-Net \citep{xia2021soe} propose a set of learnable SIFT-like filters to capture detailed local information from the point cloud data, which can be directly embedded into the PointNet network.

KPConv (Kernel Point Convolution) \citep{thomas2019kpconv} introduces a novel convolution operation that operates directly on individual points, using kernel points to define the convolution kernel. It allows adaptive receptive fields and efficient computation for encoding features from point clouds.
PF-Net \citep{huang2020pf} extracts features from the point cloud by modeling the point cloud as a fractal set, and uses a fractal convolutional neural network to process the point cloud to obtain features after layered downsampling.

Due to the high popularity of the Transformer architecture \citep{vaswani2017attention}, many researchers have incorporated it into existing frameworks for 3D feature extraction.
Studies such as \cite{zhao2021point}, \cite{guo2021pct}, \cite{pan20213d}, \cite{zhong2021point}, \cite{yu2021pointr}, \cite{xia2023casspr} and \cite{Chen_2023_CVPR} have integrated the Transformer encoder into PointNet or graph structures to extract features from point clouds.
\cite{wang2022learning} introduced a novel encoder using neighbor pooling based on the activation instead of the farthest point of an object. This structure could help to capture more meaningful structures of the objects.

Some other approaches employ voxelization techniques to convert point clouds into volumetric representations, allowing for the extraction of both local and global features through three-dimensional convolutions. 
Notable examples include VoxNet \citep{maturana2015voxnet}, Voxelnet \citep{zhou2018voxelnet}, PCNN \citep{hua2018pointwise}, GRNet \citep{xie2020grnet}, and CASSPR \citep{xia2023casspr}.
Instead of learning with all voxels of the 3D extent, \cite{yan2022shapeformer} introduced a model that can learn the completion of point clouds with a sequential representation of the voxels, which are encoded with local geometry features and voxel locations.

The Gridding Residual Network (GRNet) \citep{xie2020grnet} introduced an innovative approach for the representation of disordered point clouds. 
By transforming these point clouds into a structured 3D grid, the GRNet effectively captures and incorporates the inherent spatial structure of the point clouds, thereby enhancing the quality and accuracy of point cloud completion.

These methods derive shallow feature maps that capture boundary and shape information, offering rich local details. 
Deeper features, on the other hand, represent more abstract global characteristics. 

In this work, the adopted point cloud feature extraction involves voxelization (also referred to as gridding in the following text) of the point cloud and subsequent application of three-dimensional convolutions. 
This approach exhibits strong robustness and little susceptibility to noise.


\subsection{Point Cloud Completion}
\label{sec:Point Cloud Completion}

In recent years, there has been a growing research effort dedicated to learning-based models for point cloud completion. 
Many of these endeavors have focused on completing synthesized 3D CAD model point clouds, particularly for objects such as airplanes, vehicles, and chairs found in well-known datasets like ShapeNet \citep{chang2015shapenet}. 

Given that the point cloud completion task focuses on generating new points or surfaces, the decoder component of the network plays a critical role in this process.
The most common decoder directly generates points from latent feature maps using Multi-Layer Perceptrons, such as in PointNet, and PointNet++. 
However, such methods cannot fill large holes and missing parts. 

FoldingNet, as introduced by \cite{yang2018foldingnet}, is one of the innovative works, where the folding layer was introduced to decode features for point cloud generation. 
This method assumes that the surface of the point cloud can be folded from a two-dimensional plane, so its reconstruction network consists of two folding layers, which fold the two-dimensional plane grid into a three-dimensional point cloud. 
This component has been widely utilized in numerous subsequent studies, such as PCN \citep{yuan2018pcn}, FP-net \citep{wen2020point}, VPC-Net \citep{xia2021vpc}, PoinTr \citep{yu2021pointr}, and ASFM-Net \citep{xia2021asfm}.

Instead of folding geometry, AtlasNet, introduced by \cite{groueix2018papier}, employs a novel method for 3D shape representation. Its decoder can generate a set of parameterized surface elements, and then assemble them to form 3D mesh models. 
Furthermore, there are many other decoders, such as TopNet \citep{tchapmi2019topnet}. This decoder models the point cloud generation process as a rooted tree growth process. 
The features of child nodes are learned from parent node embeddings.

Besides 3D CAD model point clouds, there is ongoing research focusing on densifying the sparse point cloud data.
Scancomplete~\citep{dai2018scancomplete} applied fully-convolutional neural networks to complete indoor RGB-D sensor data.
By transforming incomplete signed distance functions (SDFs) into complete meshes, it effectively filled gaps in the data.
Further research utilized semantic segmentation information. 
S3CNet~\citep{cheng2021s3cnet} performs the completion task for a sparse 3D point cloud and 2D bird's-eye view jointly, using a semantic segmentation, e.g., provided by SemanticKITTI.
JS3C-Net~\citep{yan2021sparse}, on the other hand, follows a semantic segmentation-first approach, leveraging the features extracted from segmentations for the completion sub-network. 

Much of the current research in this field, as previously mentioned, focused on CAD-generated 3D object shapes, RGB-D data from indoor environments, and efforts to increase the density of sparse point cloud data gathered from rotating multi-beam LiDAR systems, such as the ones produced e.g.\ by Velodyne, Hesai, or Ouster.
LiDAR mobile mapping data has received much less attention in this context.

A common solution is to eliminate occlusion through the fusion of multiple data sources. One approach involves merging multi-temporal data from the same source \citep{
schachtschneider2017assessing} or combining different sources, such as MLS and Handheld MLS data \citep{gonzalez2024santiago}.
However, these methods also result in increased data acquisition costs due to the additional data requirements.

Existing heuristic solutions have tried to identify the occluded regions from MLS data and then employ vertical 2D projected raster images of both ground and facade to generate points for the occluded area \citep{balado2020automatic,barros2022visibility}.
However, it was noted that the generated points deviate in height from the surrounding ground surface or in depth from the facade, resulting in occasional sporadic malformations.

\cite{liu2022detection} designed a specialized workflow for handling occlusions on curbstones. 
This process begins by detecting the direction of the curb using a rule-based approach. Subsequently, the curb area was described as a geometric structure comprising two horizontal planes interrupted by a vertical plane, facilitating the use of linear interpolation to fill the gap. 
However, due to the complexity of real-world scenarios, this method may be limited in its applicability to relatively straightforward scenes. 
Therefore, employing a learning-based approach for training holds greater promise.

An additional significant hurdle involves the lack of available training data necessary for the development of the model. 
Manually generating unmeasured scenes is not feasible, further complicating the training data collection process.

There are various scanners with differences in capturing geometry and point density, necessitating a universal solution for generating training data.
The scanners all rely on a direct line-of-sight between the scan head and the object, and as long as the corresponding position of the scanner for each measurement point is known, this characteristic can be leveraged to synthesize occlusions.

Therefore, in this work, we have developed an innovative technique that allows us to generate diverse and realistic urban point cloud scenes, with and without vehicle occlusions, for model training. 
We further propose a novel Scene Gap Completion Network that leverages the strengths of both GRNet and FoldingNet.
This network is designed to achieve point cloud scene completion, specifically addressing real-world scenarios encountered in LiDAR mobile mapping data.


\begin{figure*}[h]
\centering
\includegraphics[width=.7\textwidth]{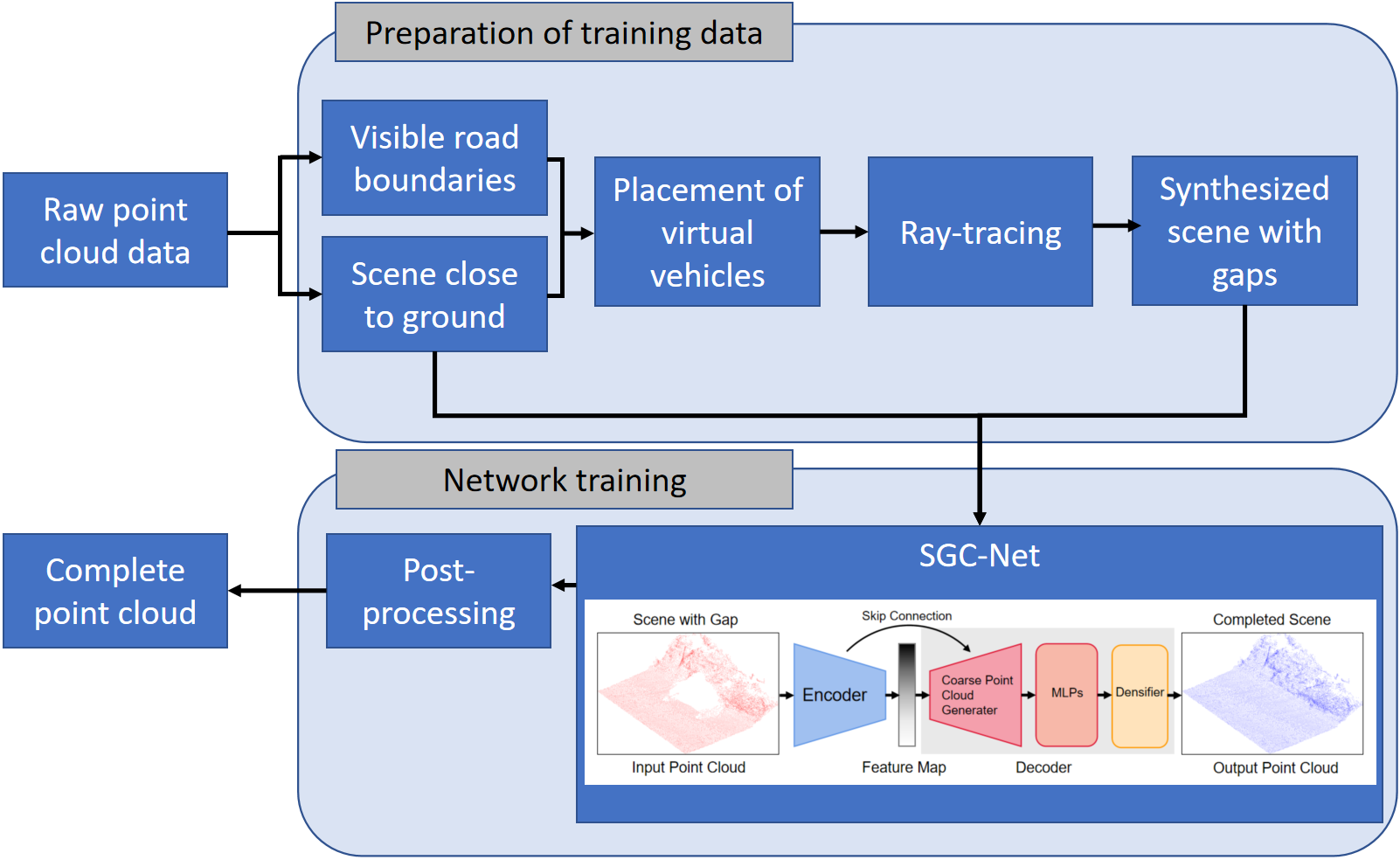}
\caption{Overview of the proposed process. It involves two main steps. The first step is preparing training data by synthesizing point cloud scenes with gaps using ray-casting. The second step uses this data to train the proposed SGC-Net model.}
\label{fig:method_overview}
\end{figure*}

\section{Methodology}
\label{sec:Methodology}

The workflow of the proposed method is illustrated in Figure \ref{fig:method_overview}. 
There are two main steps: (1)~the preparation of training datasets, and (2)~the training of neural network models. 

Since our aim is to learn a model in a supervised manner, creating a dataset for model training is a crucial task.
While our approach requires paired point clouds, one complete and one partial, which result from occlusions, it is often challenging to obtain such data. 
In some cases, multiple measurements taken at different times may be necessary to capture the scene without occlusions. 
However, this approach can be impractical and time-consuming, especially in complex or dynamic urban environments. 
To address this challenge, we synthesize the training data as explained in Section~\ref{sec:Dataset Preparation}. 
We extracted the necessary 3D scene information from the vast point cloud raw data, ensuring direct accessibility for subsequent work. 
We devised a method to locate complete scene data without vehicle occlusion for dataset production. 
Training data is obtained by resampling the scene through the ray-casting algorithm by placing virtual vehicles within the scene. 

The proposed SGC-Net model consists of an encoder module and a decoder module, wherein the decoder module is further subdivided into a coarse point cloud generation module, an MLPs module, and a densification module.
For a detailed explanation, please refer to Section~\ref{sec:Network Architecture}.

\subsection{Preparation of training data}
\label{sec:Dataset Preparation}
In order to generate the dataset for model training, two steps are involved: (1) the pre-processing of LiDAR mobile mapping data, and (2) the generation
of the training dataset.

\subsubsection{Pre-processing of LiDAR mobile mapping data} 

The LiDAR mobile mapping system used in this work is the VMX-250 from Riegl Laser Measurement Systems\footnote{Riegl VMX-250 Data Sheet. \url{http://www.riegl.com/uploads/tx_pxpriegldownloads/10_DataSheet_VMX-250_20-09-2012.pdf} (Accessed on 31.07.2023)}. 
It includes two Riegl VQ-250 laser scanners, each capable of measuring up to 300,000 points per second. The data for positions and orientations are provided by GNSS and IMU, and are further adjusted based on the reference data from SAPOS\footnote{SAPOS Service (German).\url{https://www.lgln.niedersachsen.de/startseite/geodaten_karten/geodatische_grundlagennetze/sapos/sapos-services-undbereitstellung-143814.html} (Accessed on 31.07.2023).} in Lower Saxony, Germany. 
The LiDAR mobile mapping data used in this research was captured in the city center of Hildesheim, Germany.

\begin{figure}
    \centering
    \includegraphics[clip, trim=100 150 490 140, width=.32\textwidth]{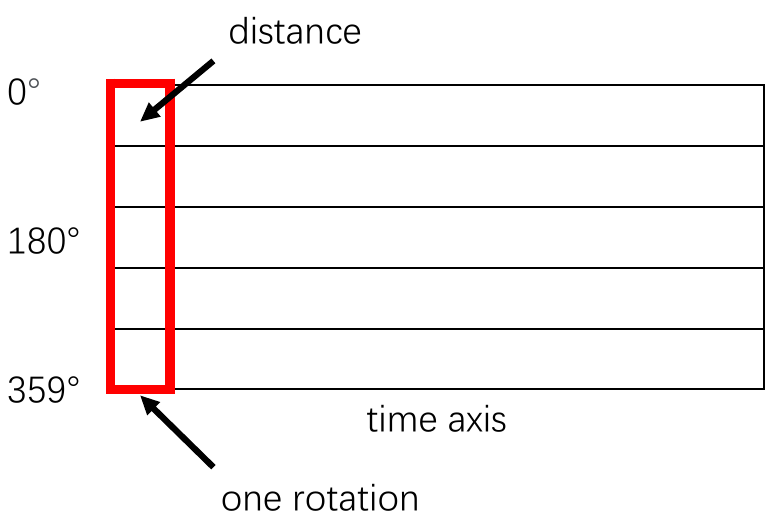}\\
    \caption{The structure of a scan strip image \citep{feng2022determination}.}
    \label{fig:scanstrip_image}
\end{figure}

\paragraph{\textbf{Alignment of LiDAR scans}}

During data capture, the LiDAR mobile mapping system acquires the same locations multiple times. This repetition, primarily due to GNSS errors, may result in substantial offsets between the point clouds across different time epochs, often in the range of several decimeters. 
As a result, it becomes necessary to first adjust scans from different epochs, particularly for large-scale LiDAR mobile mapping measurements. 
This adjustment was carried out by the method proposed by \cite{brenner2016scalable}, which applies a least squares adjustment to calculate trajectory corrections based on the 3D points observed across all epochs. This is the same as the process used in \cite{feng2022determination}

\paragraph{\textbf{LiDAR scan strips}}

Our work primarily utilizes \textit{scan strips}. They are 2D images or arrays that consist of raw measurement values in the sequence they were collected with respect to the measurement angle and time (see Figure \ref{fig:scanstrip_image}). 
Each column in the scan strip image corresponds to a full 360$^{\circ}$ 
rotation of the laser scanner mirror, with consecutive rotations occupying subsequent columns. 
With the forward motion of the platform during data collection, this array can be interpreted as a cylindrical projection of the recorded 3D points.
The number of pixels in the horizontal direction is determined by the acquisition time, while the vertical dimension comprises 3,000 pixels, signifying that the scanner captures 3,000 points per 360-degree rotation. 
Through this structure, we can store various types of information, encompassing the 3D positions of measured objects, 3D positions of measurement heads, local normals, reflectance, etc.

The key benefit of this structure is its ability to largely preserve spatial proximity, meaning that pixels in the scan strip that are close to each other correspond to 3D points in object space that are typically also nearby. 
However, certain exceptions may occur during data capture. For instance, when the scanning vehicle makes turns at intersections or encounters strong depth changes, such spatial proximity may not always hold true.

\begin{figure*}[h!]
\centering
\includegraphics[width=.75\textwidth]{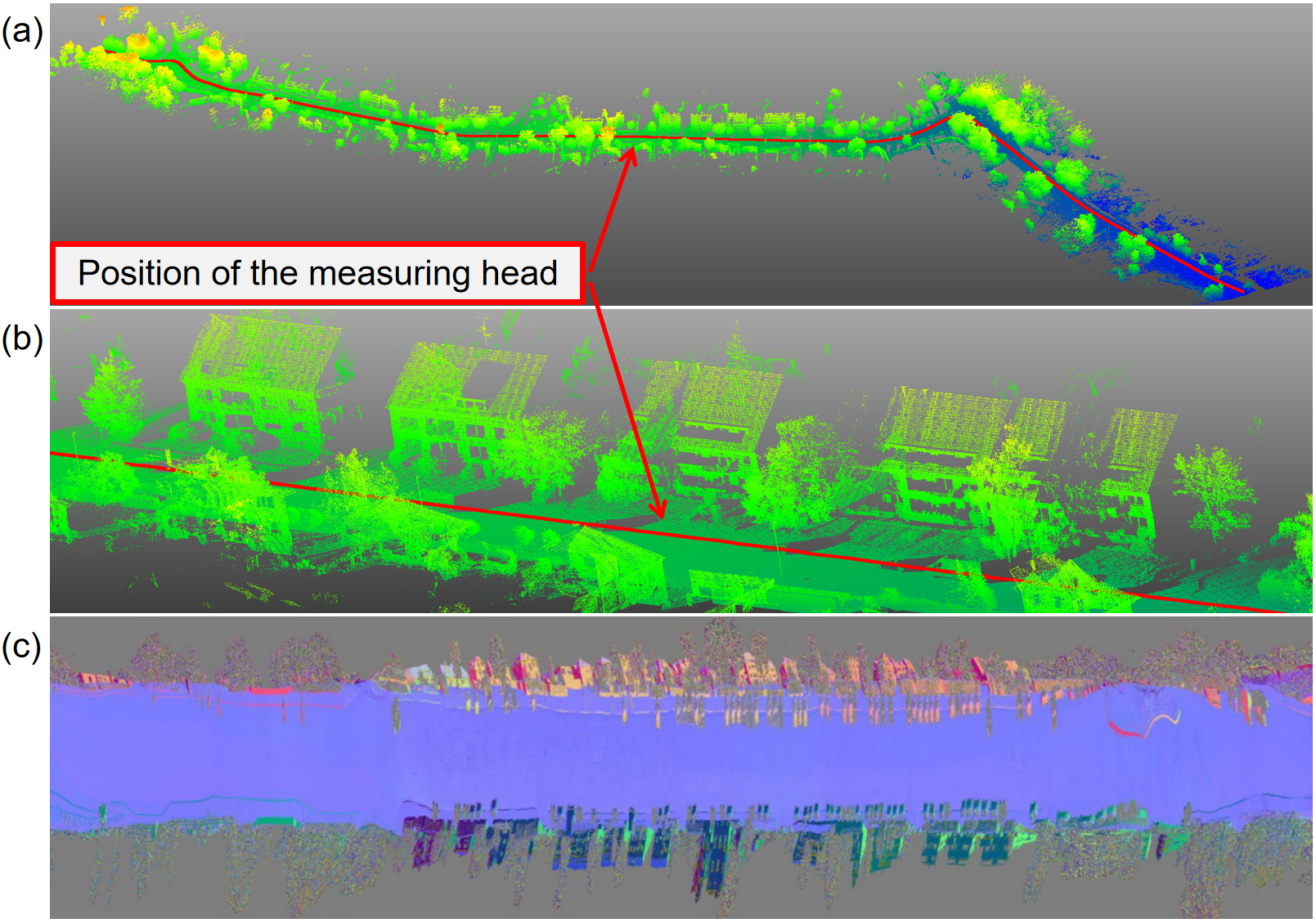}
\caption{This example showcases one scan of the original data. (a) The point clouds are represented with colors corresponding to their heights, while the red lines indicate the positions of the measuring head. (b) A closer view of the point cloud shows the locations of the measuring head. (c) One of the corresponding scan strips, with colors indicating the orientation of the normal vectors.
}
\label{fig:data_visual}
\end{figure*}
Figure \ref{fig:data_visual} illustrates the visualization of the LiDAR mobile mapping point cloud data, together with the measuring head's positions. 
Given that the mobile mapping system is equipped with a pair of scanners, each round of data collection will generate two scan strips, each corresponding to one of the two scanners. One of the corresponding scan strips is visualized in Figure \ref{fig:data_visual} (c).




\paragraph{\textbf{Scene Extraction}}

This work primarily focuses on addressing the gaps that often result from parked vehicles, which is the most common situation encountered in urban areas. 
As a result, our main area of interest is concentrated near the road, as structures behind the walls and the upper parts of buildings are not relevant for training the model.
Therefore, by eliminating less relevant data points, we achieve a dual benefit: firstly, we reduce the computation challenges, and secondly, we decrease the scene complexity, making it easier for the learning models to focus on and address the gaps caused by low objects.

\begin{figure}[h]
\centering
\includegraphics[width=0.45\textwidth]{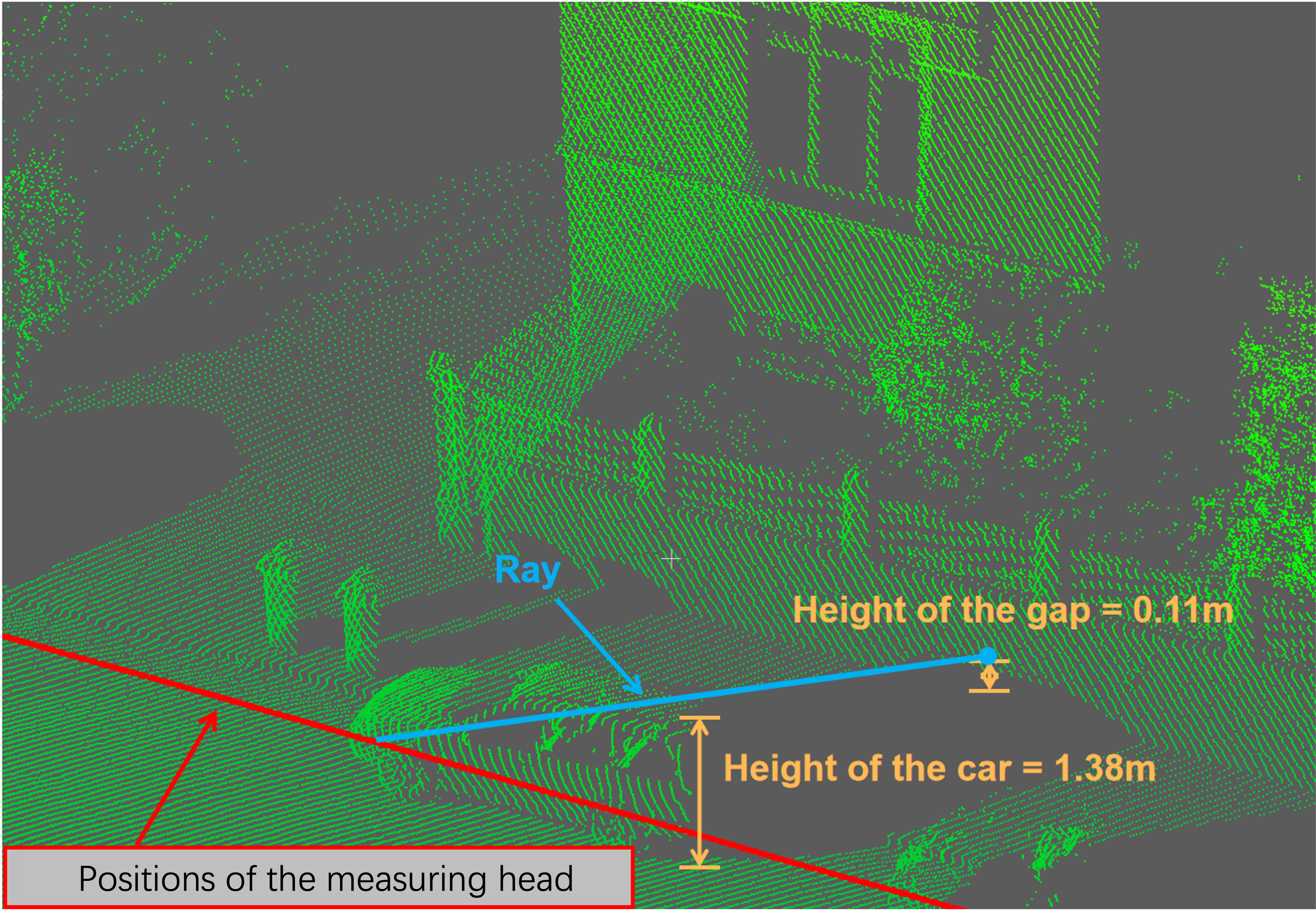}
\caption{Common gaps caused by parked cars, present both on the ground and wall. The gap height on the wall is generally lower than the height of the car, resulting in this particular gap.}
\label{fig:Wall_Gaps}
\end{figure}

We initiate the point cloud reduction process by focusing on nearby areas and limiting the data to those regions. 
Our observations indicate that, even for the widest roads in this study area, the distances between object points and the scanner head positions are predominantly within 15 meters. 
Therefore, all points
exceeding a measurement distance threshold
are excluded from the data:
\begin{equation}
\label{eq:distance threshold}
\centering{d(p_i, h_i) = \sqrt{(x_{p_i}-x_{h_i})^2+(y_{p_i}-y_{h_i})^2} \leq 15m,}
\end{equation}
where the coordinates of the measured point $p_i$ and the measuring head $h_i$ are denoted as $(x_{p_i}, y_{p_i})$ and $(x_{h_i}, y_{h_i})$, respectively, and $d(p_i, h_i)$ is their Euclidean distance.
This approach effectively filters out a significant portion of the points from inside the buildings that were scanned through the windows. 

\begin{figure*}[h!]
\centering
\includegraphics[width=.75\textwidth]{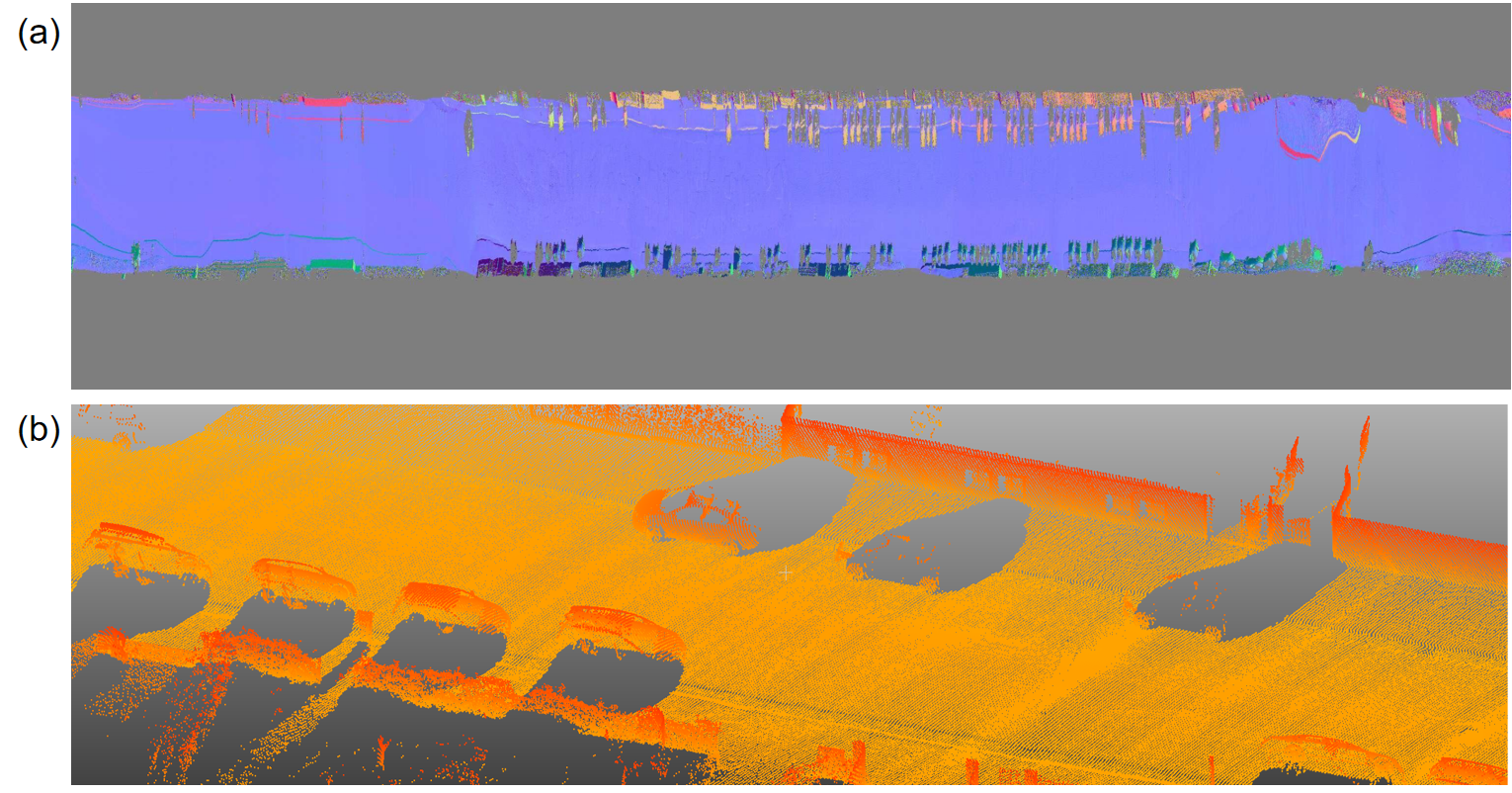}
\caption{Filtered point cloud scene. (a) Scanstrip representation of the filtered point cloud scene, with colors indicating the orientation of the normal vectors. (b) An oblique view of the filtered point cloud scene.}
\label{fig:Filtered Point Cloud Scene}
\end{figure*}

In addition to filtering based on the horizontal distance, we also consider the vertical z-axis direction during the data filtering process. 
The LiDAR measuring head is positioned on top of the mobile mapping vehicle.
The sensor head is situated at an approximate height of $H_{sensor} = 2.75$ meters above the ground.

Considering that the LiDAR measuring head is positioned more than 2 meters above ground level, and given that most vehicles have a height of less than 2 meters, wall gaps are generally under 2 meters (as illustrated in Figure \ref{fig:Wall_Gaps}).
Therefore, we adopt a filtering approach to retain only scenes within 2 meters from the ground.
Furthermore, to account for potential ground unevenness or sloping terrain, we preserve scenes up to 0.35 meters below the ground level. 
This simple
data selection ensures that we focus on the relevant areas near the road while accounting for variations in the terrain's elevation.

Consequently, we calculate the height above the ground $H(p_i, h_i)$ based on the z-coordinate of the object point ($z_{p_i}$), the z-coordinate of the measuring head position ($z_{h_i}$), and the fixed sensor height $H_{sensor}$:
\begin{equation}
\label{eq:distance_threshold1}
\centering
H(p_i, h_i) = z_{p_i} - z_{h_i} + H_{sensor}.
\end{equation}
We select scenes that exhibit a relative height between 0.35 meters below and 2 meters above ground, using:
\begin{equation}
\label{eq:distance_threshold2}
\centering
-0.35m < H(p_i, h_i) < 2m.
\end{equation}
An example of the filtered scan strip and its corresponding point cloud is illustrated in Figure \ref{fig:Filtered Point Cloud Scene}.






\paragraph{\textbf{Extraction of road boundary}}

\begin{figure}[h!]
\centering
\includegraphics[width=0.48\textwidth]{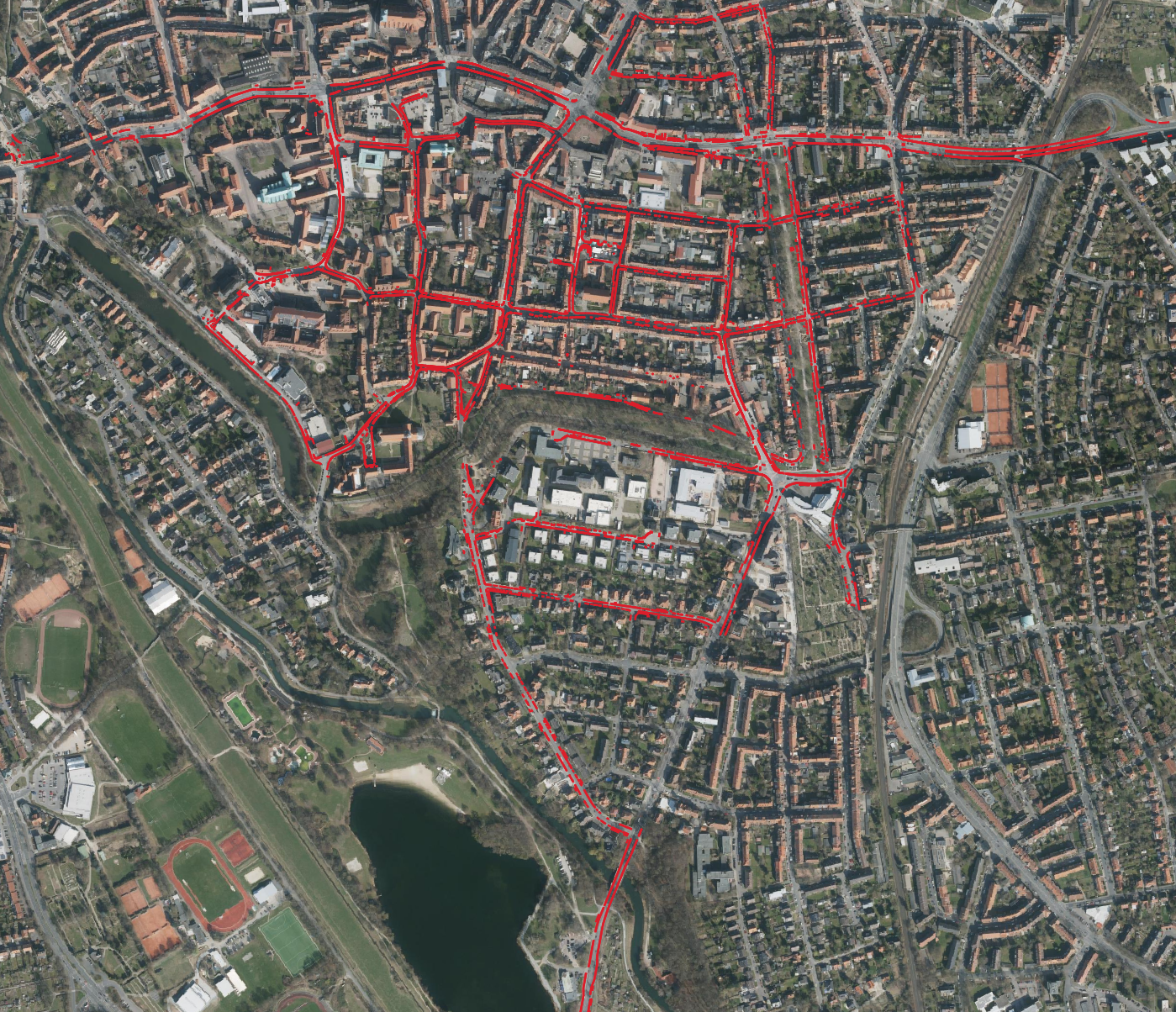}
\caption{Results of the road boundary extraction. (Basemap: Digital Orthophoto from LGLN OpenGeoData\protect\footnotemark)}
\label{fig:road_boundaries_map}
\end{figure}
\footnotetext{OpenGeoData provided by the State Office for Geoinformation and Land Surveying Lower Saxony (LGLN): \url{https://opengeodata.lgln.niedersachsen.de} (accessed on 11.11.2023)}

Given that the vehicles causing occlusion are primarily parked on the side of the road, obtaining the position of the road boundary becomes essential. 
Once we have determined the positions of the road boundaries, we can strategically place virtual vehicles to simulate the occlusion effect caused by parked vehicles on the side of the road in the point cloud data. 

Based on the normal vectors calculated from the scan strips, we further applied a region-growing algorithm to extract segments roughly perpendicular to the ground. 
This approach is similar to the building facade extraction method utilized in \cite{feng2022determination}. 
The only difference is that we select the road boundary segments according to their shape and location.

The common height of the curbstone is around 30 cm or lower. 
To depict this characteristic, we rasterize each segment with a cell size of 0.33 meters from the bird's-eye view and compute the median of the height differences within each cell. 
The median value of heights needs to be less than 30cm.
Furthermore, the segment's shape must be elongated, and we enforce a length-to-height ratio of at least five times when working with this raster image.
Lastly, they should be near the ground surface, so that the segment needs to be $\pm 0.5m$ around the ground surface.

With the implementation of all three rules, we are able to create a map highlighting the most prominent curbstones as shown in Figure \ref{fig:road_boundaries_map}.
The red lines primarily represent the positions of road boundaries. 
The interruptions in continuity within the red lines are a result of occlusions caused by vehicles or situations
with lowered curbstones at intersections or driveways.

\subsubsection{Generation of training dataset with ray-casting} \label{sec:DatasetGeneration}
To facilitate the training of the model with point cloud data, we need to standardize the input.
This involves placing car models in locations free from occlusions. 
By employing ray-casting, we produce distinctive \textit{point cloud scenes} with and without gaps. These scenes consist of square regions of point clouds, centered at the car's position. We regarded the scene before ray-casting as a complete scene, while the one after is considered as a scene with gaps.

With all the above information as input, the dataset for model training is prepared with the following three steps: (1)~Selection of proper locations for placing car models, (2)~Creation of point cloud scenes with occlusions through ray-casting, and (3)~Preparation of the dataset.

\paragraph{\textbf{(1) Selection of proper locations for placing car models}}

With the map of road boundaries as potential locations for placing virtual vehicles to simulate occlusions, there are many candidates where we can place them. 
However, not all of them are suitable, especially when we need to avoid densely parked areas.

Therefore, two thresholds are set to ensure reasonable scene selection. 
The first threshold requires the length of the visible road boundary to be over 7 meters, accommodating most private cars and considering reduced parking space due to a curved road.
The second threshold ensures that selected locations are not too close to the endpoints of the curb stone line segment, with a minimum distance of 3.5 meters to guarantee suitability for parking a vehicle.


We employ ShapeNet vehicle models \citep{chang2015shapenet} as the foundation for placing virtual vehicles within the comprehensive point cloud scene. The ShapeNet dataset contains an extensive collection of 3,533 car mesh models. From this pool, we carefully 
select 1,500 models, excluding those deemed impractical for road scenarios, such as excessively elongated presidential cars and high-performance racing cars. These chosen models closely resemble standard vehicles and are randomly positioned within the complete scene to facilitate dataset generation.
Ensuring the placement of virtual vehicles adheres to three pivotal considerations: 

\begin{enumerate}
  \item Virtual vehicles must be situated in positions that conform to traffic regulations and are logically sound within the scene.
  
  \item The vehicle's chassis must seamlessly align with the terrain, even on varying slopes or uneven surfaces.
  
  \item The vehicle's orientation should align correctly with the proper direction, typically parallel to the road boundary.
\end{enumerate}

\begin{figure}[h!]
\centering
\includegraphics[width=.49\textwidth, trim=0 0 0 60, clip]{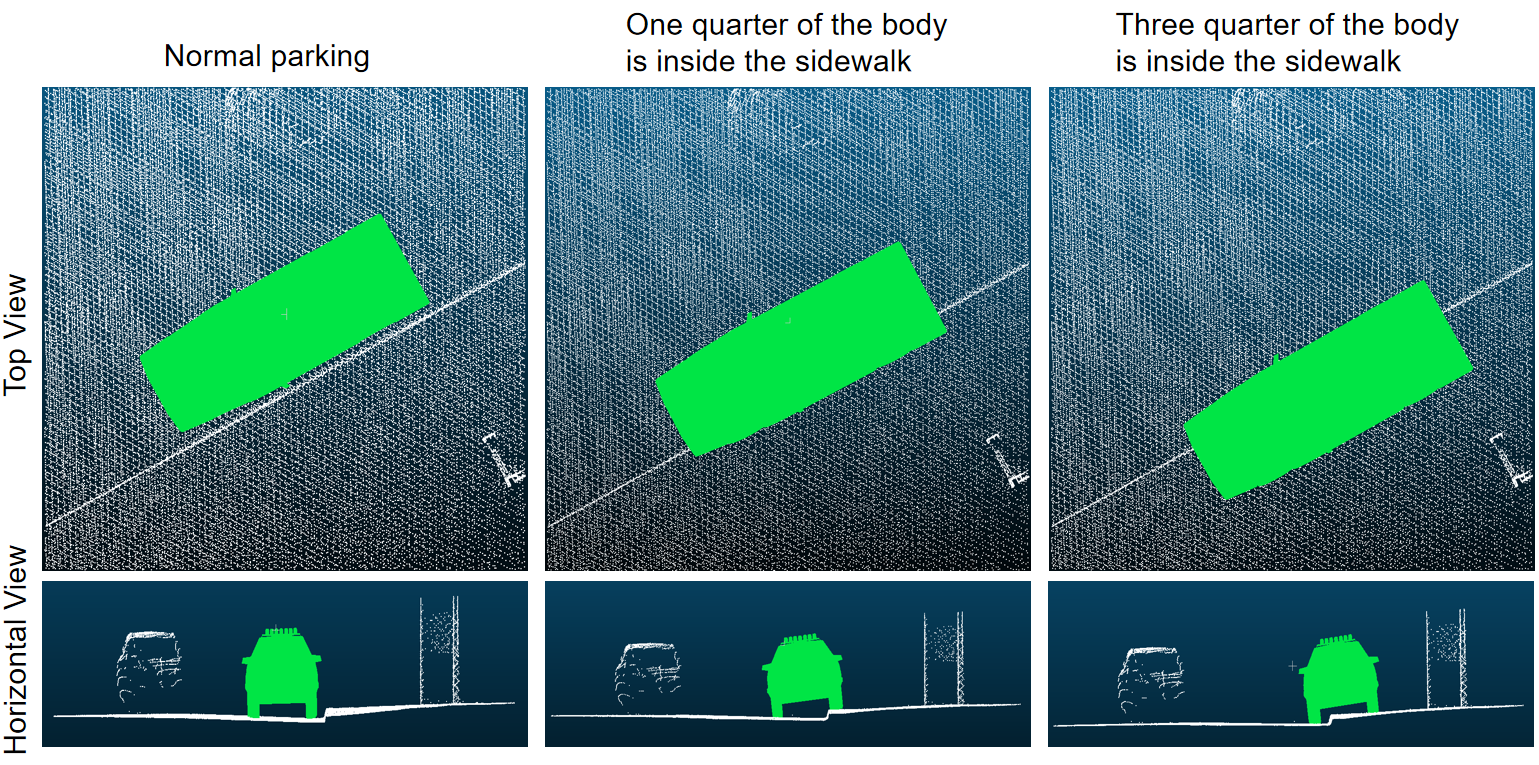}
\caption{Various parking methods along roads: entire car on the road (left), 1/4 car width on the sidewalk (middle), and 3/4 car width on the sidewalk (right).}
\label{fig:Different Ways of Parking}
\end{figure}

\noindent To meet these considerations, we follow a three-step process:

\begin{enumerate}
  \item  \textbf{Location Adjustment}: The initial location points obtained in the previous step are situated on the road boundary. However, real-world vehicles are often parked slightly away from the boundary. 
  To simulate the vehicle's location accurately, we shift it by half the vehicle width and introduce a random distance of less than 0.3 meters as the offset between the vehicle and the road boundary. 
  The virtual vehicle is placed at that point.
  
  \vspace{2mm}

  Additionally, we observe that on some narrow roads, vehicles are parked differently, with a portion of the body on the sidewalk and parked parallel to the road boundary. For this scenario, we employ a random function that positions the car body randomly, ranging from a quarter to three-quarters of the car's width inside the sidewalk, as shown in Figure \ref{fig:Different Ways of Parking}.
  \item \textbf{Alignment to the ground surface}: With the data points retrieved from the previous step within a 2-meter radius, RANSAC fitting was applied to obtain the ground plane parameters, allowing us to determine the precise z-axis height at the location point. 
  Subsequently, we rotated the car to align with this plane and accurately positioned it at the appropriate height and location.
  
  \item \textbf{Vehicle Orientation}: In the final step, we aligned the vehicle's lateral to the previously determined road direction.
  Occasionally, we placed some cars perpendicular to the road direction.
\end{enumerate}


By addressing these three aspects, we can proficiently position virtual vehicles within the complete scene, generating a point cloud that replicates real-world parking, encompassing a variety of occlusions and parking scenarios.

\paragraph{\textbf{(2) Creation of point cloud scene with occlusions through ray-casting}}



We then need to crop the point cloud scene with the car in the center of the scene to create individual scene input.
However, having an overly precise center may reduce the model's robustness to adapt to variations in position. 
Therefore, we introduce randomness and noise during cropping to create a more flexible center, enabling the model to better handle different car positions. 
We select a random center within a 0.2-meter radius of the actual car location to obtain the cropped scene as a model input. 
This approach simulates real-world variations, enhancing the model's robustness.

\begin{figure}[t]
\centering
\includegraphics[width=.45\textwidth]{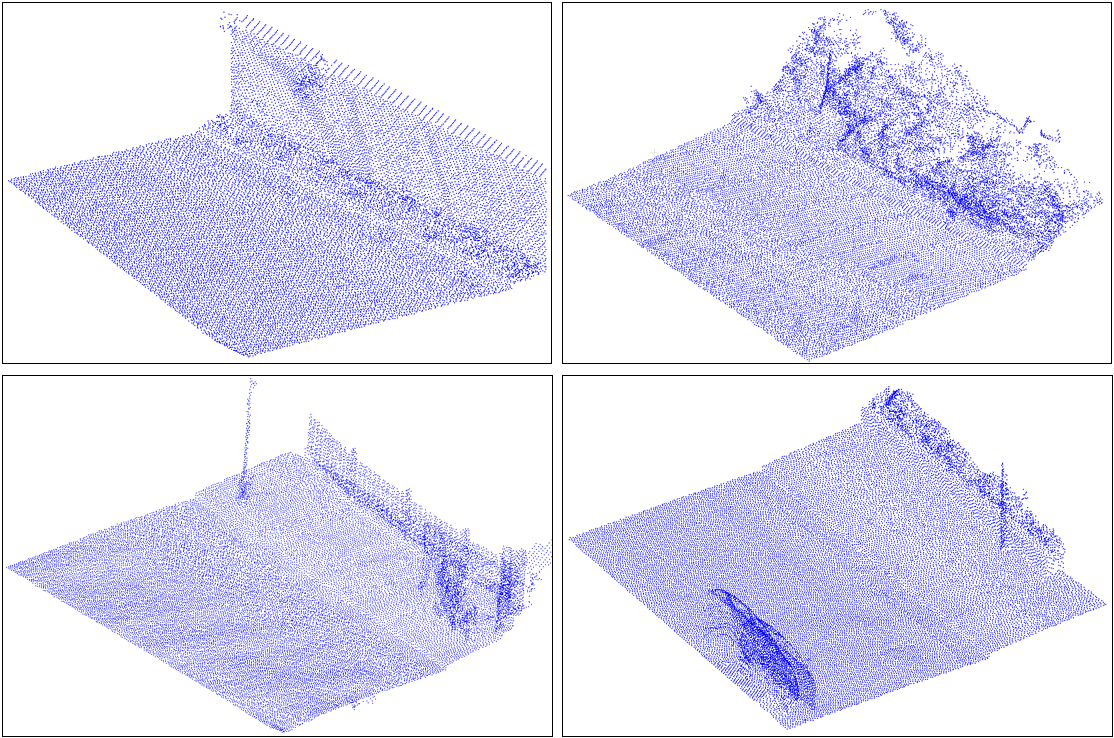}
\caption{Examples of complete scenes that were cut out.}
\label{fig:complete_scene}
\end{figure}

Based on the filtered point cloud scene, we cut out square blocks of 8m $\times$ 8m from the horizontal plane as the complete scene data (examples visualized in Figure~\ref{fig:complete_scene}). 

\begin{figure}[h!]
\centering
\includegraphics[width=.45\textwidth]{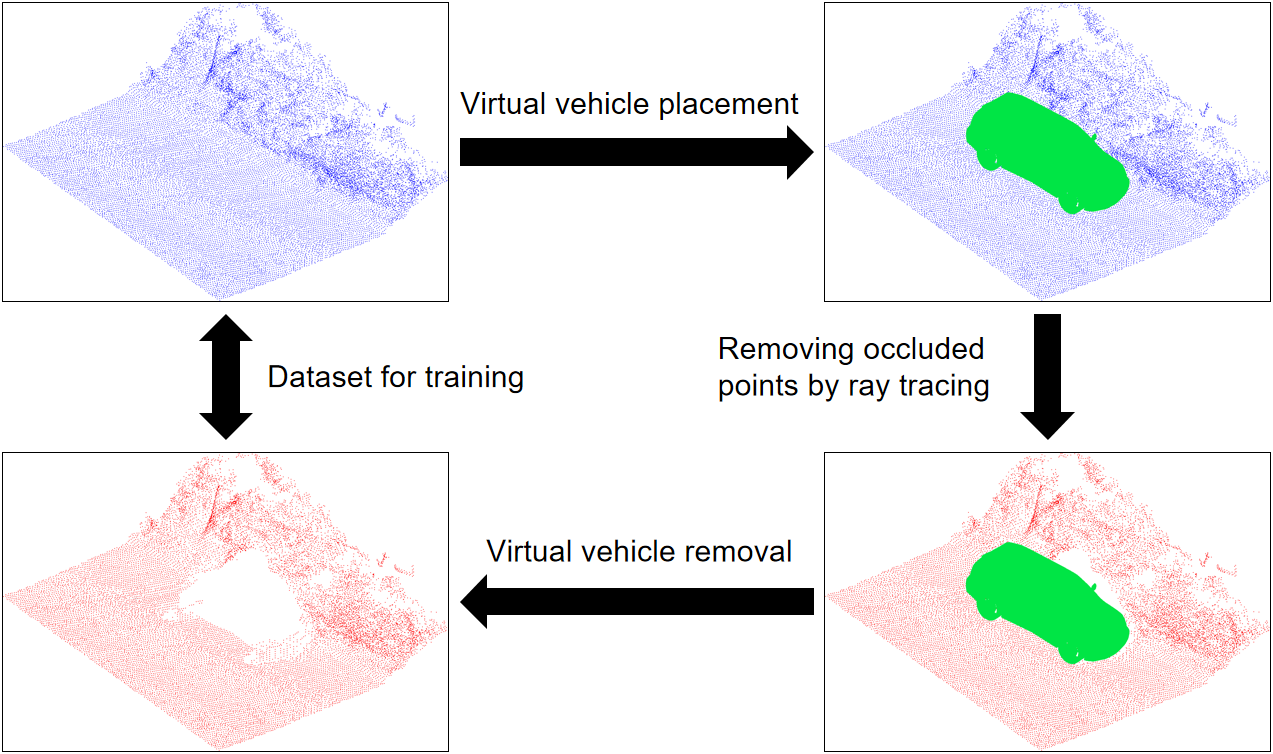}
\caption{The workflow of creating a point cloud scene with occlusions through ray-casting. In the original point cloud scene (top left), a car model is placed on the road (top right). Subsequently, laser rays intersecting with this model are identified (bottom right) and then removed from the original point cloud (bottom left).}
\label{fig:Flowchart of dataset generation}
\end{figure}

After placing the virtual vehicle model within each scene, we need to identify which laser rays will be obstructed by this model, thereby causing occlusion.
Each laser ray, originating from the vehicle's head positions and targeting the corresponding points in the point cloud scene, is recorded, allowing us to reconstruct all laser rays passing through the scene.
By assessing whether laser rays intersect with the virtual vehicle model, we identify the points obscured by the vehicle and subsequently remove them from the point cloud dataset. 
The remaining points form a point cloud scene with a gap, which serves as input data for training and testing neural network models.
Figure \ref{fig:Flowchart of dataset generation} presents a flowchart, depicting this process of creating a scene with gaps by ray casting.

\paragraph{\textbf{(3) Preparation of the dataset}}

Each point cloud scene contains a substantial number of data points, and their quantities vary, posing a significant challenge for computing resources and standard neural network models, that expect a consistent input size.
Due to hardware limitations, there is a need to strike a balance between richer information and the number of data points.
Therefore, in our generated dataset, random
subsampling is applied to ensure an equal number of points in each scene.

The GPU used for this research is an Nvidia GeForce RTX 3090 Ti, providing access to 24GB of graphics memory.
After empirical experiments with various output sizes, it was determined that 27,648 points were to be sampled from the complete scene. 
This number (27~$\times$~1,024) was chosen to ensure efficient training without compromising too much on the ability to represent details.
On average, there are 4.32 points for every $10cm*10cm$ horizontal area. 
These quantities are sufficient to adequately represent the context and three-dimensional shape of an 8m $\times$ 8m square scene.
Furthermore, it is observed that the gap generally occupies 1/3 of the squared scene, thus 18,500 points are sampled from the point cloud scenes with gaps.

\subsection{Network Architecture} 
\label{sec:Network Architecture}

\begin{figure*}[h!]
\centering
\includegraphics[width=.8\textwidth]{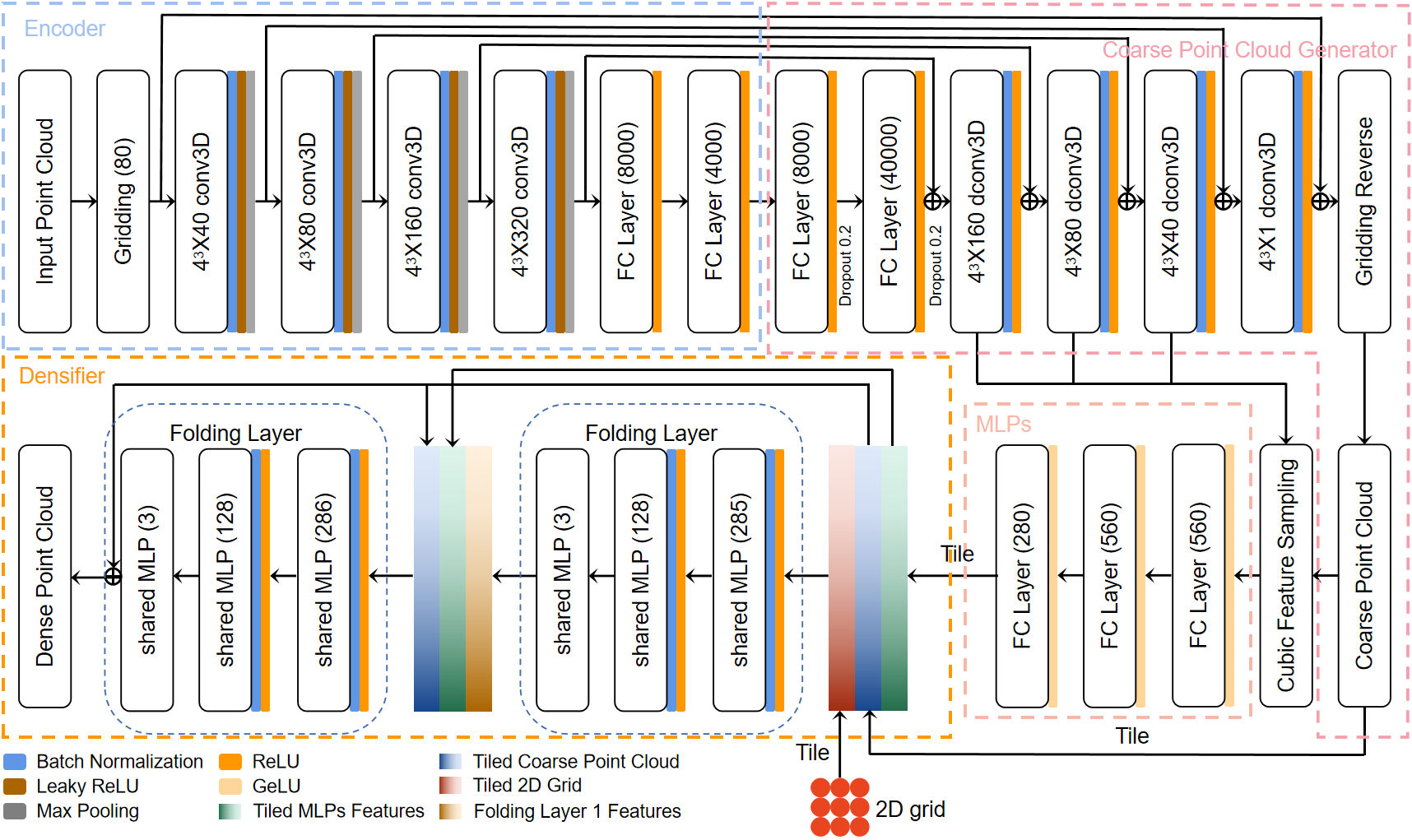}
\caption{The network architecture of SGC-Net. First, the encoder extracts global and local features from the original point cloud scene with gaps. Second, the decoder generates the completed point cloud through three components: a coarse point cloud generation module, an MLPs module, and a densification module.}
\label{fig:The Network Architecture of SGC-Net}
\end{figure*}

The network architecture of the SGC-Net is shown in Figure \ref{fig:The Network Architecture of SGC-Net}. 
It consists of two sub-networks: an encoder for feature extraction, and a decoder for reconstructing a coarse and a dense point cloud. 
The design of our network is primarily inspired by the encoder part of GRNet \citep{xie2020grnet} and the decoder part of FoldingNet \citep{yang2018foldingnet}.
Unlike point cloud completion for 3D shapes, where the focus is on generating object details with reasonable protrusions, in urban scenes, we prioritize obtaining smooth surfaces such as ground and walls, as well as areas with curbs. 

FoldingNet is widely used as a decoder in many point cloud completion networks, including PCN \citep{yuan2018pcn} and PoinTr \citep{yu2021pointr}. 
It involves folding the 2D grid into 3D space and then refining the point cloud by further folding it in 3D space. This process enables more effective construction of the complete structure of the point cloud.
GRNet has presented superior encoding capabilities by directly applying convolutions on point cloud data while preserving their structure and contextual information. 
Therefore, we propose integrating FoldingNet with GRNet's encoder to potentially enhance performance. 
The effectiveness of combining these two components is validated through an ablation study in Section \ref{sec:ablation}.

First, the encoder (Sec.~\ref{sec:Encoder}) extracts global and local features from the original point cloud scene with gaps. 
Second, the decoder (Sec.~\ref{sec:Decoder}) generates the completed point cloud through three components: a \textit{coarse point cloud generation module} that creates a sparse, coarse point cloud by leveraging information from the global feature map and the local feature maps; an \textit{MLPs module} that generates more dense point clouds with finer structures by reducing the dimension of cube features; and a \textit{densification module} that combines the feature outputs by the MLPs, coarse point cloud coordinates, and the 2D grid to generate the final dense point cloud through two folding layers. 
Each part is described in detail in the following subsections.

In this approach, we utilize the 3D grid representation of point clouds along with 3D convolution and transposed convolution operations.
Our main objective is to enhance the learning of contextual relationships within the 3D geometric structure.
This representation allows the network to capture the local geometric structure and maintain superior spatial information compared to methods like PCN, which rely on point-implicit features for shape reconstruction. 
In the decoder, we utilize a FoldingNet-based approach to learn local features more effectively compared to a fully connected layer \citep{yuan2018pcn}.

\subsubsection{Encoder} \label{sec:Encoder}
The encoder $E$ aims to extract a series of local feature maps $F_{local}$ ranging from shallow to deep, as well as a global feature map $F_{global}$. 
This enables the decoder to estimate the missing portions of the 3D point cloud scene. Since 3D scenes encompass a wealth of edge and shape information, it is crucial for the encoder to extract local features from the scene with gaps, allowing the decoder to perform 3D coordinate regression for dense point cloud generation.

Within the encoding process, the point cloud is initially transformed into an $80 \times 80 \times 80$ 3D grid using GRNet's gridding method \citep{xie2020grnet}, and the values within each grid cell are computed. 
These grid values resemble grayscale values in an image and can have their features learned by convolutional neural networks. 
As illustrated in Figure \ref{fig:The Network Architecture of SGC-Net}, the encoder comprises four 3D convolutional layers, followed by two fully connected layers, which are used to obtain the global features $F_{global}$. Each convolutional layer includes a bank of $4^3$ filters with padding of 2. This is then followed by batch normalization, Leaky ReLU activation, and a max pooling layer using a kernel size of $2^3$. The four 3D convolutions have 40, 80, 160, and 320 convolution kernels, which can progressively abstract the feature map $F_{local}$ from shallow to deep, with deeper features being more abstract. Shallow features can estimate the geometric structure of the point cloud, while abstract features aid in estimating the shape, density, and arrangement of the point cloud. Ultimately, the global features $F_{global}$ are abstracted by two fully connected layers with dimensions of 8000 and 4000. These features can assist in estimating the complete point cloud scene within the decoder. The parameter weights of the encoder $E$ is denoted as $w_{E}$, and the encoder can be defined as:
\begin{equation}
\label{eq:encoder}
\centering{[F_{global}, F_{local}]=E(P_{input} \mid w_{E})}
\end{equation}
where $P_{input}$ is the point cloud scene with gaps.

\subsubsection{Decoder} \label{sec:Decoder}
The decoder $D$ is designed to combine global features $F_{global}$ and local features $F_{local}$ in order to generate dense, homogeneous, and complete 3D point cloud scenes. The decoding process is divided into three functionally distinct parts: the coarse point cloud generation module $D_{coarse}$, which generates the basic 3D structure and clear boundaries of the point cloud scene; the Multilayer Perceptrons module $D_{MLPs}$, which learns the coarse point cloud based on its 3D grid features to obtain the implicit features of each point for subsequent densification; and the densification module $D_{densification}$, which increases the density of the point cloud, learns local features based on the MLPs features and the location information of the coarse points, and generates a clear 3D structure of the scene. The decoder is defined as follows:
\begin{equation}
\label{eq:decoder}
\centering
\begin{split}
P_{output}=D (F_{global}, F_{local} \mid w_D) \\ D=D_{densification} \circ D_{MLPs} \circ D_{coarse}
\end{split}
\end{equation}
%
where $P_{output}$ is the output dense point cloud, $w_D$ denotes the weight parameters of $D$. 
These three modules are introduced below.

\paragraph{Coarse Point Cloud Generation Module} \label{sec:Coarse Point Cloud Generation Module}
The coarse point cloud generation module, denoted as $D_{coarse}$, is composed of two fully connected layers with dimensions 8000 and 40000, and it also includes four 3D transposed convolutional layers. 

In the first step, the global feature $F_{global}$ is expanded from 4,000 to 40,000 dimensions with the initial two fully connected layers, which are accompanied by a ReLU activation and a dropout operation with a 0.2 probability.

Then, it is reshaped into grids measuring $320 \times 5^3$ and fed into these four following 3D transposed convolutional layers.
The channels of the 3D transposed convolutional layers are 160, 80, 40, and 1, respectively. Each layer has a bank of $4^3$ filters with padding of 2 and stride of 1, followed by a batch normalization layer and a ReLU activation.
Skip connections are employed, where the local feature $F_{local}$ is seamlessly integrated with the decoder part of the same size. This facilitates the direct pass of edge information into the decoder, enabling the reconstruction of finer details in the output.



In the final step, the last layer of 3D transposed convolution produces an $80^3$ grid, which generates up to 493,039 ($79^3$) points through the Gridding Reverse introduced in GRNet \citep{xie2020grnet}.
Random sampling is employed to select points, and ultimately, 3,072 points form the resulting coarse point cloud, denoted as $P_{coarse}$. 
These points are randomly sampled, and 3,072 points are output as a coarse point cloud $P_{coarse}$. 
The Gridding Reverse process, however, doesn't generate points when all eight vertices of a cell are 0. Consequently, the number of points generated by Gridding Reverse is notably less.
Based on our experimentation, this selection of 3,072 points proves sufficient for adequately representing the scene's shape and its fundamental boundary structure. Beyond this point count, increasing accuracy is constrained, while the hardware resources required for the computation significantly escalate.

With $w_{D_{coarse}}$ denoting the weight parameters of the coarse point cloud generation module, the coarse point cloud generation module can be defined as:
\begin{equation}
\label{eq:coarse point cloud generation module}
\centering{P_{coarse}=D_{coarse}(F_{local}, F_{global} \mid w_{D_{coarse}})}
\end{equation}

\paragraph{MLPs Module}\label{sec:MLPs Module}
The MLPs module $D_{MLPs}$ is composed of three Cubic Feature Sampling operations as introduced in \citep{xie2020grnet} and three fully connected layers with dimensions of 560, 560, and 280, respectively. 
Each fully connected layer is followed by a GeLU activation. 

First, the three Cubic Feature Sampling operations extract the point features $F_{cubic}$ of the coarse point cloud from the feature map of the 3D transposed convolution. 
Three Cubic Feature Samplings extract for each point 2,240 features (i.e., $8 \times (160+80+40)$) from the feature maps of their 8 surrounding vertices after applying each 3D transposed convolutional layers.
These features are then passed through three layers of fully connected layers to obtain the dimensionality-reduced feature $F_{MLPs}$ for each point. The dimensionality of the features is 280. The purpose of designing MLP is to prepare for the subsequent densification of the point cloud. 


With $w_{D_{MLPs}}$ denoting the weight parameters of the MLPs module, the MLPs module is defined as:
%
\begin{align}
    F_{cubic} &= Cubic\_Feature\_Sampling(P_{coarse}) \label{eq:MLPs Module1} \\
    F_{MLPs} &= D_{MLPs}(F_{cubic} \mid w_{D_{MLPs}}) \label{eq:MLPs Module2}
\end{align}
The role of the MLPs module is to generate implicit MLPs' features for each point so that the subsequent densification module can better learn local features and improve the detailed structure of the output point cloud.

\paragraph{Densification Module}\label{sec:Densification Module}
The function of the densification module is to increase the density of the point cloud and to learn local details better with fewer parameters than fully-connected decoder by outputting $r$ times the number of points of the coarse point cloud $N_{P_{coarse}}$, $N_{P_{output}}=rN_{P_{coarse}}$ \citep{yuan2018pcn}. 
The densification module is composed of folding-based decoders, which are good at predicting and approximating smooth surfaces that represent the local geometry of a shape \citep{yuan2018pcn}. 
The Densification Module is composed of two folding layers, and each folding layer consists of three layers of shared multi-layer perceptrons (shared MLPs) with dimensions [285, 128, 3] and [286, 128, 3]. 
A batch normalization layer and a ReLU activation follow after the first two shared MLPs in each folding layer.

In the first stage, we tile the points in $P_{coarse}$ to produce a dense point set with a number of $P'_{coarse}:=rN_{P_{coarse}} \times 3$. 
Then, we tile the MLPs feature $F_{MLPs} :=N_{P_{coarse}} \times 280$ of each point to generate the dense point tiled feature $F'_{MLPs} :=rN_{P_{coarse}} \times 280$ corresponding to the dense tiled point set in the same way. 
Third, we created a $u \times u$ 
squared grid ($u^2=r$) centered at the origin and transformed the coordinates of the points into a $r\times2$ matrix, and the matrix is then tiled $N_{P_{coarse}}$ times into a $rN_{P_{coarse}} \times 2$ tiled matrix. 

In the second stage, the tiled point set, tiled feature, and tiled matrix are concatenated and sent to the first folding layer with a total of 285  dimensions (280 MLPs features + 3 points' coordinates + 2 grids' coordinates). 
Then, the output of the first folding layer (3 dimensions) is concatenated with the tiled point set and tiled feature, then the tiled matrix total of 286 dimensions (280 MLPs features + 3 points' coordinates + 3 output features of the first folding layer) is sent to the second folding layer to generate an $M$ matrix with size $M:=N_{P_{output}} \times 3$. This can be seen as a non-linear mapping that transforms the 2D grid into a smooth 2D manifold in 3D space \citep{yuan2018pcn}. 

Finally, to generate the output point cloud $P_{output}:=N_{P_{output}} \times 3=rN_{P_{coarse}} \times 3$, the coordinates of each point in the tiled coarse point cloud $P_{coarse}$ are added to the matrix $M$. 

With $w_{D_{densification}}$ denoted as the weight parameters of the densification module, this module can be defined as:
\begin{equation}
\label{eq:densification Module2}
\centering{P_{output}=D_{densification}(F_{MLPs} \mid w_{D_{densification}})+P'_{coarse}}
\end{equation}

\subsection{Loss Function} \label{sec:Loss Function}
The loss function measures the difference between the generated point cloud and the ground truth reference point cloud, thus enabling network optimization.
Since point clouds are three-dimensional data composed of disordered points, this function must be uncorrelated with the permutation of the points while calculating the topological distance. 
In this context, the Chamfer Distance (CD) can serve as a proper loss function for measuring the dissimilarity, as demonstrated in \cite{fan2017point}.

The CD between the output generated point cloud $P_{output}$ and the ground truth $P_{gt}$ is defined as:
\begin{equation}
\label{eq:cd}
\begin{split}
d_{CD}(P_{output}, P_{gt})=&\frac{1}{n_{P_{output}}} \sum_{x \in P_{output}} \min_{y \in P_{gt}} {\parallel x-y \parallel}^2_2\\
&+\frac{1}{n_{P_{gt}}} \sum_{y \in P_{gt}} \min_{x \in P_{output}} {\parallel y-x \parallel}^2_2
\end{split}
\end{equation}
where $n_{P_{output}}$ and $n_{P_{gt}}$ are the total number of points in the output point cloud and ground truth point cloud, respectively.
CD calculates the average squared nearest neighbor distance between the output point cloud $P_{output}$ and the ground truth point cloud $P_{gt}$. We use a symmetric CD definition. The first term in Eq. \ref{eq:cd} is the average value of each point in the output point cloud to the point of the minimum distance in the ground truth point cloud, and the second term is vice versa. Notably, $n_{P_{output}}$ and $n_{P_{gt}}$ do not need to be identical to calculate the CD.

The network employs a gridding operation to convert point clouds into grids, and the coarse point cloud generated by the decoder is also in the form of grids.
As a result, we employ the Gridding Loss, a concept introduced in GRNet by \citep{xie2020grnet}, which leverages L1 loss for grid-based representations, much similar to images. The Gridding Loss can be formally defined as follows: 


\begin{equation}
\label{eq:Gridding Loss}
\centering{L_{Gridding}(P_{pred}, P_{gt})=\frac{1}{N^3_G} \sum \parallel W^{output}-W^{gt} \parallel}
\end{equation}
where $W^{pred}$ are the grid values generated by the 3D CNN, $W^{gt}$ are the values of the input point cloud directly after gridding, and $N^3_G$ is the number of grids.
We have referenced the training strategy of GRNet\footnote{Github repository of GRNet. Source: \url{https://github.com/hzxie/GRNet}}.
Initially, the network focuses on optimizing the raw point cloud through the minimization of the CD error until the loss function converges (regarded as Stage 1). Subsequently, we fine-tune the grid to enhance finer details using the Gridding Loss  (regarded as Stage 2). 
This strategy could better capture the underlying geometric structure, as declared by the GRNet authors.


Our loss function $L$ is to use Gridding Loss or CD to optimize the loss of coarse point clouds $P_{coarse}$, and to use CD to optimize the loss of output point clouds $P_{output}$. 
The overall loss function can be defined as the following two stages. \\
During Stage 1, we use:
\begin{equation}
\label{eq:Loss_stage1}
\begin{split}
L(P_{coarse}, P_{output}, P_{gt}) &=d_{CD}(P_{coarse}, P_{gt})\\ 
&+\alpha d_{CD}(P_{output}, P_{gt})
\end{split}
\end{equation}
During Stage 2, we use:
\begin{equation}
\label{eq:Loss_stage2}
\begin{split}
L(P_{coarse}, P_{output}, P_{gt}) &=L_{Gridding}(P_{coarse}, P_{gt})\\ 
&+\alpha d_{CD}(P_{output}, P_{gt})
\end{split}
\end{equation}
where $\alpha$ is a hyperparameter that is used to balance the weights between the losses of each component. 


\subsection{Post-processing}\label{sec:Post-processing Method}
The point cloud generated by the network, denoted as $P_{output}=\{y_i: i=1,...,N'\}$, contains more points than the input point cloud $P_{input}=\{x_i: i=1,...,N\}, N<N'$, since the gaps are filled in this process. 
It is important to note that the input point cloud is not a subset of the output point cloud $P_{input} \notin P_{output}$. 
The input point cloud contains real collected data, providing more accurate information about the actual 3D scene compared to the generated point cloud. 
In order to preserve the utmost precision of original point cloud data, only the gap parts of the newly generated point cloud are cropped and merged with the input data with gaps to post-process the output.

To identify the gap regions, we utilize a k-d tree spatial data structure to identify points within the generated point cloud that lie within a proximity of less than 8cm from the points present in the input point cloud containing gaps. 
Points within this range are considered to have overlaps with the input data, where the exact measured points from the input data are used as the final output.
Regarding this hyperparameter, we can only set this threshold empirically. A larger threshold would remove more points that could be informative, while a smaller threshold would retain much more noise. 

\section{Experiments}\label{cp:experiment}

In this section, we introduce the experimental data, evaluation metrics, and baseline methods used to assess the performance of point cloud generation methods. 
Subsequently, we present the experimental details, showcasing the performance of the proposed SGC-Net for point cloud scene completion. 
Lastly, we summarize and discuss the experimental results.

\subsection{Experimental Dataset}\label{sec:Experimental dataset}

\begin{figure}[t]
\centering
\includegraphics[width=0.45\textwidth]{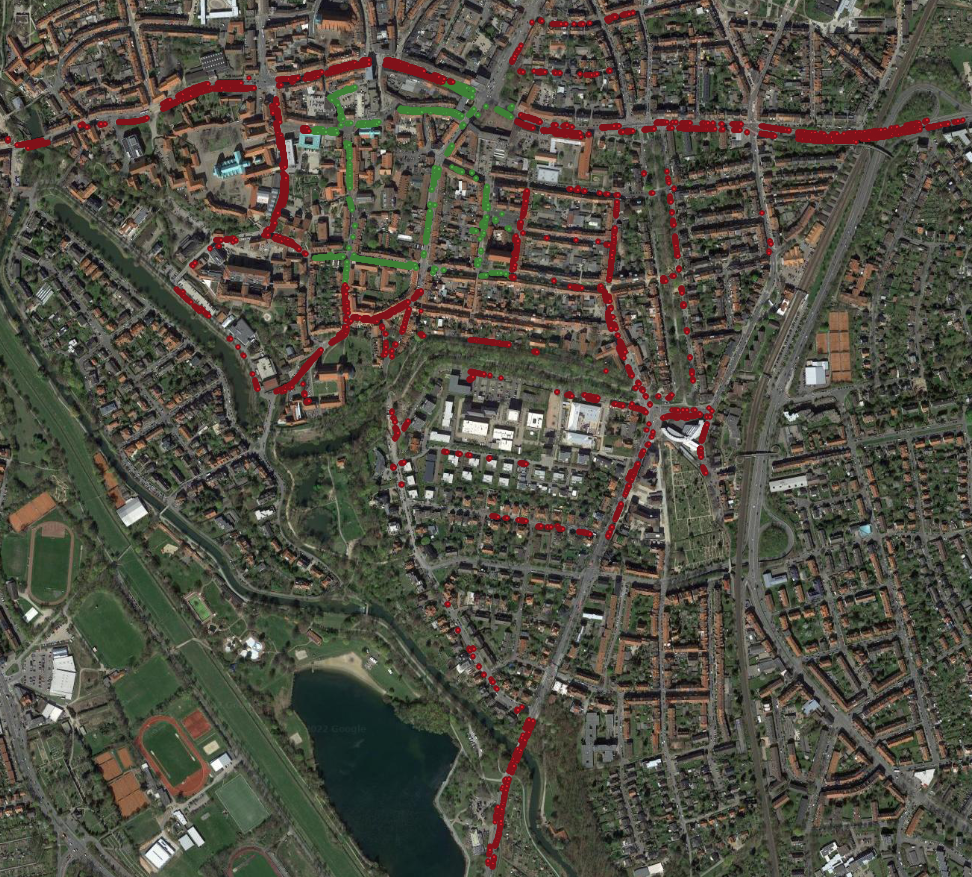}
\caption{Overview of the locations from which gap-free scenes were cropped: the locations of red points are used for training and green for testing. (Basemap: Digital Orthophoto from LGLN OpenGeoData)}
\label{fig:Distribution of the Dataset}
\end{figure}

Following the steps presented in Section \ref{sec:Dataset Preparation}, we extracted a total of 13,786 point cloud scenes and combined them into a dataset. 
They are evenly distributed along the road boundaries in the city. 
Figure \ref{fig:Distribution of the Dataset} provides an overview of the locations from which gap-free scenes were cropped. The locations of red points are used for training and green for testing.
The test data and training data are geographically separated, which is to ensure that scenes from the test data are not present in the training data. 
With this split, the test set contains 2,513 scenes, and the training set contains 10,973 scenes.

\begin{table}[h]
    \centering
    \caption{Number of training data, test data and validation data in the dataset.}
    \label{table:Dataset}
    \begin{tabular*}{\tblwidth}{@{}LCCC@{}}
    \toprule
        Dataset & Train set & Test set & Validation set \\ \midrule
        Number of scenes & 10,973 & 2,513 & 300 \\ 
    \bottomrule
    \end{tabular*}
\end{table}

Moreover, we expanded the value by a factor of 3 along the z-axis direction of height, as the height difference in the vertical direction of road boundaries is relatively small and insignificant. The choice of this operation is further investigated in an ablation study in Section \ref{sec:ablation}. 
The three dimensions of the point cloud data are each normalized to the range of [-1, 1]. This preprocessing step contributes to enhancing the performance and stability of the deep neural network during training.

\begin{figure*}[h!]
\centering
\includegraphics[width=.9\textwidth]{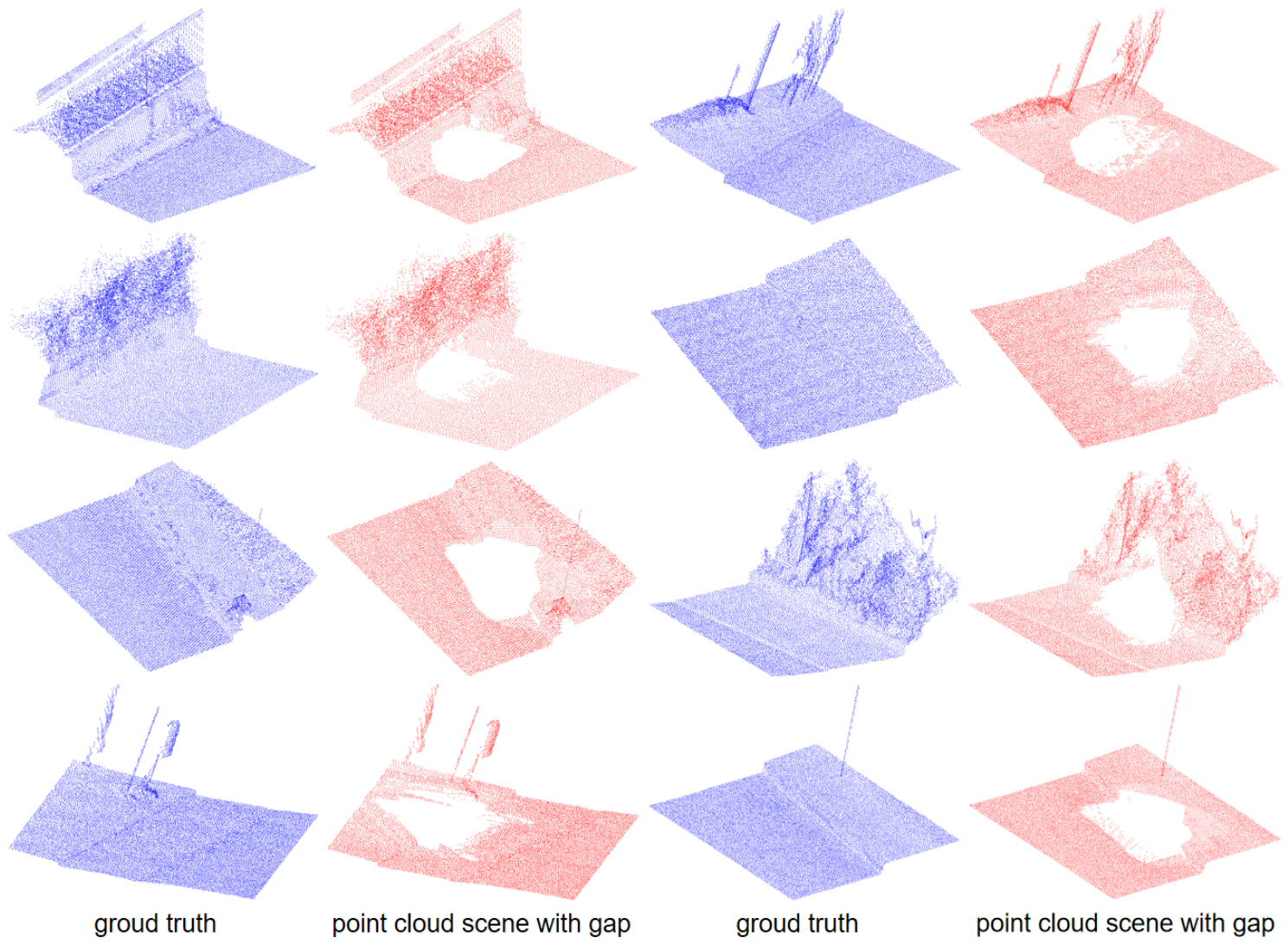}
\caption{Examples from the training dataset. The point clouds visualized with blue color are the target ground truth scene, and the point clouds in red are the scenes with gaps.}
\label{fig:DatasetExamples}
\end{figure*}
\begin{table*}[h]
    \centering
    \caption{Quantitative comparison of point completion results of our method with the state-of-the-art baseline methods. The values in the table are the mean values of CD, F-Score, Precision and Recall, calculated over all point cloud scenes in the test set.}
    \label{table:Point Completion Results}
    \begin{tabular*}{\tblwidth}{@{}LCCCC@{}}
    \toprule
        \textbf{Methods} & \textbf{CD $\downarrow$}  & \textbf{F-Score $\uparrow$} & \textbf{Precision $\uparrow$} & \textbf{Recall $\uparrow$}\\ 
         & \textbf{($\times 10^{-4}$)} &  &  &  \\ \midrule
        PCN & 11.077 & 0.732 & 0.501 & 0.569 \\
        GRNet (Gridding $64^3$) & 1.969 & 0.720 & 0.679 & 0.766  \\
        GRNet (Gridding $80^3$) & 1.878 & 0.751 & 0.712 & 0.794  \\
        SGC-Net (without post-processing) & 1.604 & 0.768 & 0.732 & 0.808  \\
    \midrule
        PoinTr  & 1.283 & 0.866 & 0.850 & 0.884  \\
        PoinTr (with post-processing) & 1.269  & 0.867 &  0.851 &  0.884 \\
        SGC-Net (with post-processing) & \textbf{1.129} & \textbf{0.892} & \textbf{0.856} & \textbf{0.932}  \\
    \bottomrule
    \end{tabular*}
\end{table*}

\begin{figure*}[h]
\centering
\includegraphics[width=.95\textwidth]{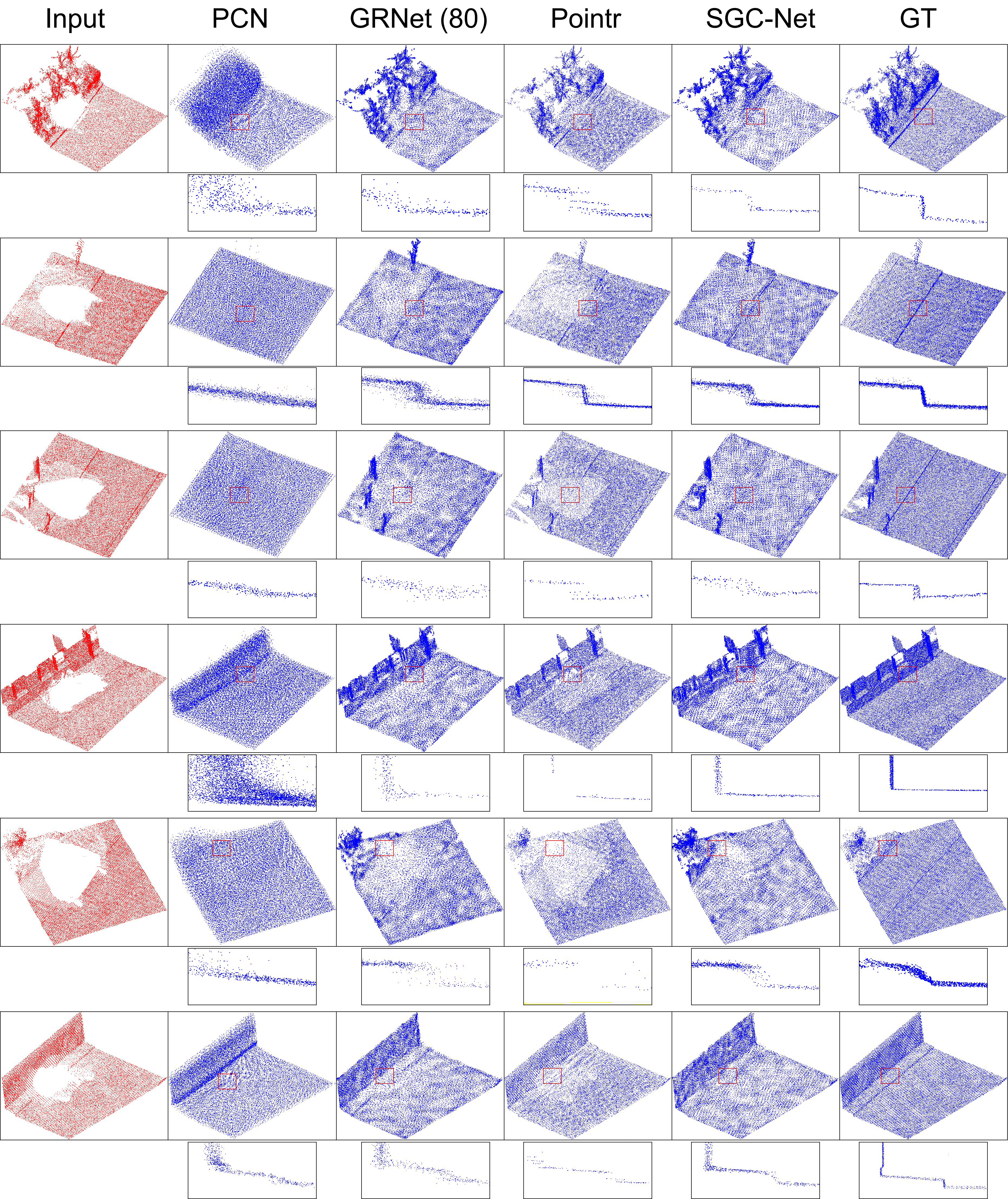}
\caption{Qualitative comparison of the completion results with different methods for the test set. The red boxes highlight specific areas for comparison, and the corresponding profiles are displayed below each scene. GT represents the ground truth of the 3D scene.}
\label{fig:Qualitative Completion Results}
\end{figure*}

We then perform data augmentation on the training data: during training, scenes are randomly rotated around the origin z-axis by [0\textdegree, 90\textdegree, 180\textdegree, 270\textdegree], followed by random flipping along the x-axis or y-axis after rotation.

Furthermore, we used the same method to generate another 300 scenes to serve as the validation set (non-overlapping with either the training set or test set), allowing us to evaluate training results during the training process and save the best model. 
In summary, Table \ref{table:Dataset} displays the dataset's composition, while Figure \ref{fig:DatasetExamples} presents several examples of the dataset.

\subsection{Evaluation Metrics}\label{sec:Evaluation Metrics}
Two metrics are used to evaluate our point cloud scene completion method between the completed point cloud and the ground truth. 

The most commonly applied metric is the Chamfer Distance (CD) (Eq. \ref{eq:cd}). 
The definition and calculation method have been given in Section \ref{sec:Loss Function}. The second metric we applied is the F-Score. 
As suggested by \cite{tatarchenko2019single}, the F-Score more explicitly evaluates the distance between the surfaces of the objects, which is calculated as the percentage of points that was reconstructed correctly. The definition of F-Score is as follows:
\begin{equation}
\label{eq:F-score}
\centering{F{\text -}Score(d)=\frac{2P(d)R(d)}{P(d)+R(d)}}
\end{equation}
where $P(d)$ and $R(d)$ are the precision and recall at distance threshold $d$ respectively. 
According to the work from \cite{tatarchenko2019single}, the distance threshold cannot exceed 1\% of the total length, and we take the value 0.01 (corresponds to 4cm in real scale before normalization) in this work. 
The definition of precision and recall is:
\begin{equation}
\label{eq:precision}
\centering{P(d)=\frac{1}{n_{output}}\sum_{y \in P_{output}}(\min_{x \in P_{gt}} \parallel x-y \parallel < d)}
\end{equation}
\begin{equation}
\label{eq:recall}
\centering{R(d)=\frac{1}{n_{gt}}\sum_{y \in P_{gt}}(\min_{x \in P_{output}} \parallel x-y \parallel < d)}
\end{equation}
where $n_{gt}$ and $n_{output}$ are the number of ground truth point clouds $P_{gt}$ and output point clouds $P_{output}$, respectively.

\begin{figure*}[h]
\centering
\includegraphics[width=.95\textwidth]{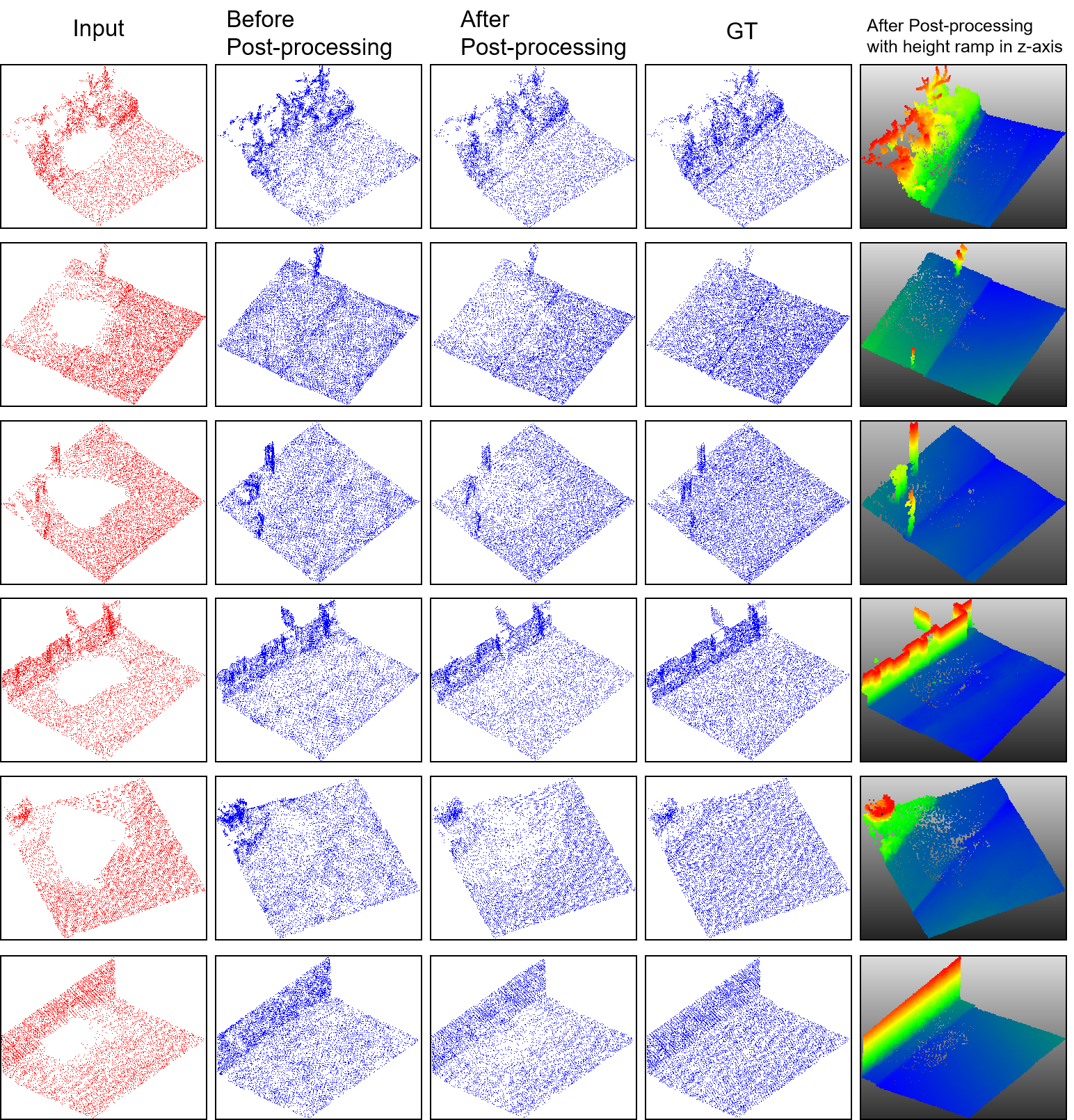}
\caption{Qualitative comparison before and after post-processing of the point cloud scene completed by SCG-Net. The post-processed point cloud is color-coded to represent height variations along the z-axis, transitioning from blue to red for lower to higher heights. This color scheme enhances the visualization of shape details in the generated point cloud.}
\label{fig:Post-Processing}
\end{figure*}

\subsection{Implementation Details}\label{sec:Implementation Details}
We implemented our SGC-Net, using PyTorch, and trained it on an NVIDIA 3090Ti GPU with 24GB of memory. 
All models are optimized with the Adam optimizer, with parameters set to $\beta_1=0.9$ and $\beta_2=0.999$. 
During training, the batch size is 8, the initial learning rate is 0.0002, and the learning rate decays by a factor of 0.97 every epoch. The parameters $\alpha$, used to balance each loss function in Eq. \ref{eq:Loss_stage1} and \ref{eq:Loss_stage2}, are initially set to 0.01 and are increased by a factor of 10 after every 20 epochs until their value reaches 1. 
We alter the loss function of the coarse point cloud from CD error (Eq. \ref{eq:Loss_stage1}) to Gridding Loss (Eq. \ref{eq:Loss_stage2}) after 80,000 steps to fine-tune the grid parts.




We observed that during training, when generating dense point clouds, the model gradually mitigates the effects of noise produced by the coarse point clouds. The point cloud output from the refiner progressively reduces the training loss and validation loss as the number of training steps increases, eventually converging after another 10,000 steps (i.e., step 2 in Eq. \ref{eq:Loss_stage2}).


\subsection{Evaluation of the point completion method on test set}
To evaluate the proposed method for completing point cloud gaps, the following state-of-the-art methods are used to demonstrate the performance of our model, including PCN \citep{yuan2018pcn}, GRNet \citep{xie2020grnet}, PoinTr \citep{yu2021pointr}. 
For fairness, we use the same training strategy for training and testing on the same dataset. 
The size of the output point cloud for the baseline methods and ground truth is set to 27,648 points. 
Due to the varying number of points produced by these baseline methods, it was necessary to increase the dimensions of their latter hidden layers and the output layer to achieve an output size consistent with the expected ground truth size. 

Quantitative and qualitative comparisons are presented in Table \ref{table:Point Completion Results} and Figure \ref{fig:Qualitative Completion Results}, respectively.
As can be seen from Table \ref{table:Point Completion Results}, our proposed SGC-Net method outperforms other methods. In the table, the CD error is scaled up by $10^4$ times, and the F-Score, Precision, and Recall are presented as percentages.

\begin{figure}[t]
\centering
\includegraphics[width=0.48\textwidth]{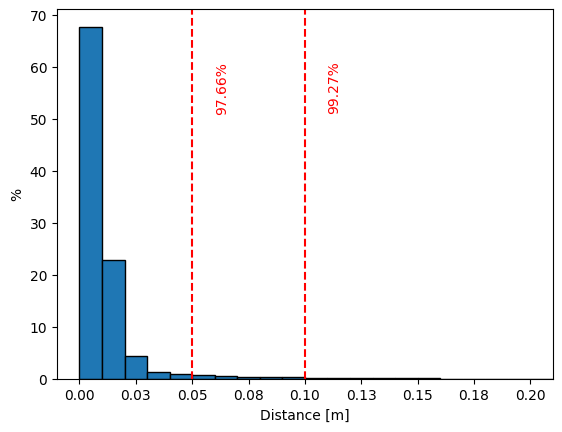}
\caption{Distribution of distances from each fill point to an estimated plane based on the nearest 15 points in the high-density point cloud.}
\label{fig:distance_histogram}
\end{figure}

The baseline methods can be categorized into two groups. The first category involves directly generating point clouds without utilizing any input point cloud as output, e.g., PCN, GRNet. These methods are comparable to our SGC-Net basic model without post-processing.
The second category involves combining the generated points with the input points that have gaps to create the output point cloud, as demonstrated in methods like PoinTr. 
This method is comparable to our SGC-Net after post-processing.

\paragraph{Group 1:}
As our model's backbone is built on GRNet, our initial step is to compare its performance with our proposed SGC-Net. 
While GRNet partitions the point cloud into $64^3$ grids, SGC-Net adopts a finer division into $80^3$ grids to achieve more detailed representations.
We trained GRNet using both $64^3$ and $80^3$ grids, with the latter yielding better results.
In comparison to GRNet (gridding $80^3$), SGC-Net before post-processing exhibits a 14.6\%, 1.5\%, 1.9\%, and 1.1\% improvement in CD, F-Score, Precision, and Recall performance metrics, respectively. 

The F-Score indicates that both GRNet (gridding $80^3$) and SGC-Net can accurately capture the scene shape, while the significant improvement in CD metric demonstrates our model's superior handling of fine structures, including boundary clarity and point cloud surface smoothness.
This enhancement is particularly significant because CD is considered more suitable for measuring the fine-grained structure of objects \citep{fan2017point}.

\paragraph{Group 2:}
Given that metrics require the calculation of the closest point distances between the output point cloud and the ground truth point cloud, it is reasonable to observe that PoinTr outperforms all baseline methods in Group 1. 
This is because PoinTr only completes the gap parts of the point cloud scene and then merges the generated points with the input points that have gaps to form the output point cloud. 
This is similar and also comparable to our method after the post-processing (as introduced in Section \ref{sec:Post-processing Method}). 

Through post-processing, areas in the scene that do not require completion are replaced with points from the input scene. 
Although our post-processing method may not entirely filter out points that do not entirely belong to the gaps, the result is smooth and seamlessly integrates with the input scene.

After post-processing, the CD error of the point cloud generated by SGC-Net decreases by 29.6\%, and the F-Score, Precision, and Recall improve by 12.4\%, 12.4\%, and 12.4\%, respectively, which is better than the performance of other models, and better than the second best PoinTr in CD, F-Score, Precision, and Recall's performance metrics by 12.0\%, 2.6\%, 0.6\%, and 4.8\%, respectively. To better demonstrate the impact of post-processing, as a comparison, we applied the same post-processing to the output of PoinTr. The results indicate minimal differences.

More qualitative results are presented in Figure \ref{fig:Qualitative Completion Results} and \ref{fig:Post-Processing}. 
Figure \ref{fig:Qualitative Completion Results} reveals that SGC-Net produces smoother surfaces and sharper boundaries compared to other methods. 
As for this task, PCN can only approximate the scene's rough shape, while GRNet (gridding $80^3$), PoinTr, and SGC-Net can predict the correct geometric information.
GRNet (gridding $80^3$) produces relatively rough surfaces on flat regions.
PoinTr produces a smooth surface on the horizontal plane but struggles with vertical plane completion, resulting in unclear shape boundaries. 
Our proposed SGC-Net achieves a smooth surface and well-defined shape boundaries at the same time.
Alongside the analysis of the ground surface, further examination involving vertical planes is presented in Section \ref{sec:complete_detail}, providing comprehensive details.

Figure \ref{fig:Post-Processing} compares point clouds before and after post-processing, demonstrating that the post-processed scene effectively retains the input part's point cloud with details.
Combining the points generated only at the gap parts with the input scene yields a seamlessly integrated complete point cloud scene. 
We colored the post-processed point cloud, in the direction of the z-axis, transitioning from blue to red. 
This color variation allows us to clearly visualize the differences in elevation along the geometric boundaries, especially the boundary between the road and the sidewalk. 
From the figure, we can see that geometric boundaries can be efficiently generated using our method.

For improved interpretability of our results, we computed the distances from each filled point to an estimated plane based on the nearest 15 points in the high-density point cloud and visualized the distribution using a histogram (see Figure \ref{fig:distance_histogram}). 
Our analysis demonstrates that 97.66\% of the filled points lie within the range of 5 centimeters and 99.27\% within the range of 10 centimeters to the ground truth, indicating the ability of our method to reconstruct the surface precisely.

\subsection{Visualization of completion details}
\label{sec:complete_detail}

\begin{figure*}[h!]
\centering
\includegraphics[width=.8\textwidth]{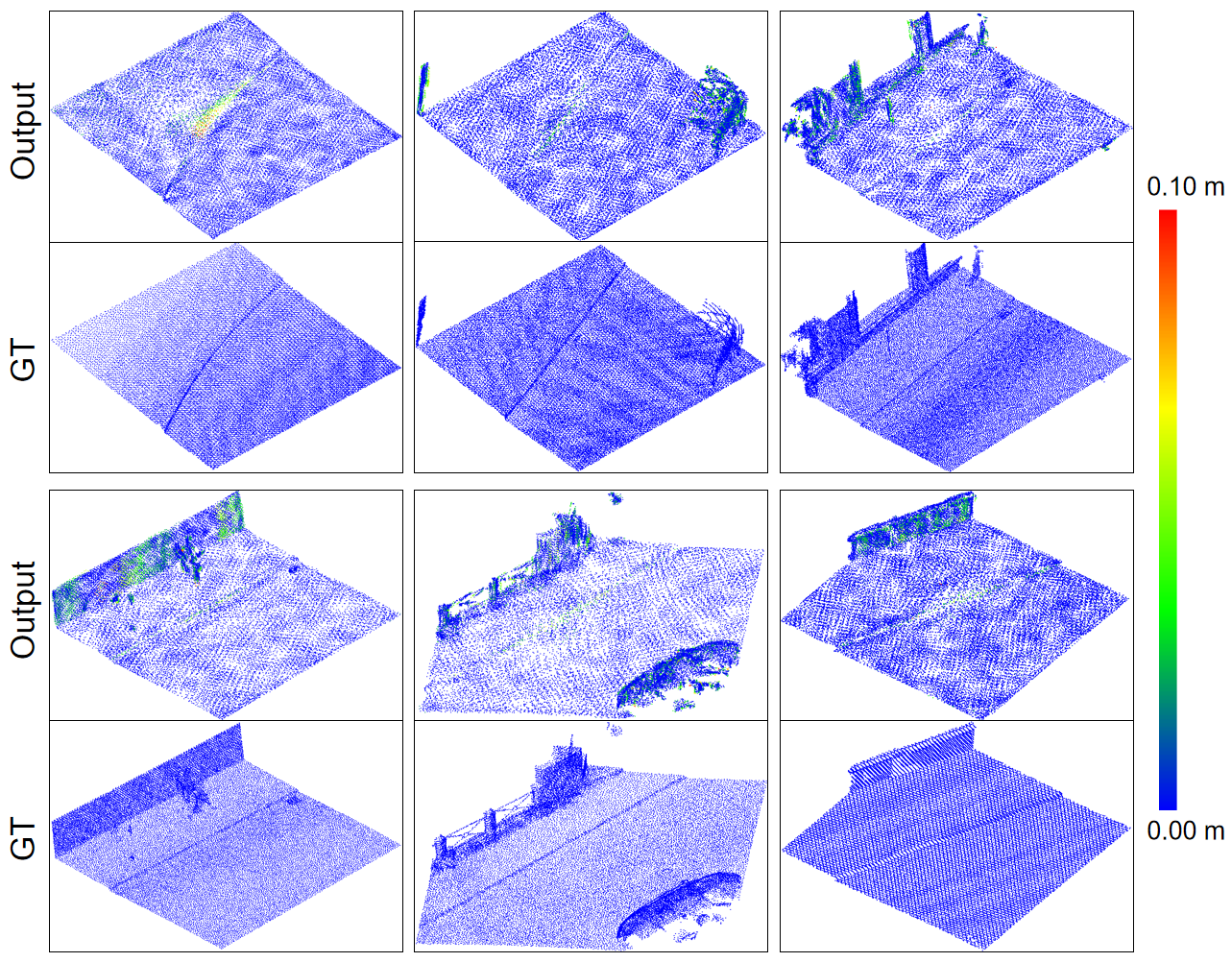}
\caption{Point cloud scene completion details, with colors indicating the normalized distance error between the generated points and their corresponding ground truth point cloud.}
\label{fig:Details of Output}
\end{figure*}

To present a more comprehensive point cloud scene completion performance, we conducted a thorough analysis by calculating and visualizing the spatial dissimilarity between each point in the generated point cloud and its closest point in the ground truth reference point cloud.

In Figure \ref{fig:Details of Output}, we present six scenes showcasing the direct model output from the test set, meaning the results are shown without any additional post-processing.
The color in the figure shows the normalized distance error between the generated points and their corresponding ground truth point cloud.
Points with a distance from the reference cloud exceeding 0.1 meters are marked in red, indicating significant generation errors. 
Differences within this range are considered acceptable, ensuring the continuity of the completed scene surface.
From the figure, it can be observed that most locations with noticeable differences are marked in green, while there is only very few locations marked in red, 
indicating that the majority of generated points fall within an acceptable range.
The points generated on flat surfaces are well-distributed and exhibit minimal errors. The areas with significant errors are primarily complex structures and road boundaries. 




\subsection{Robustness to noisy objects}
\begin{figure*}[h!]
\centering
\includegraphics[width=0.75\textwidth]{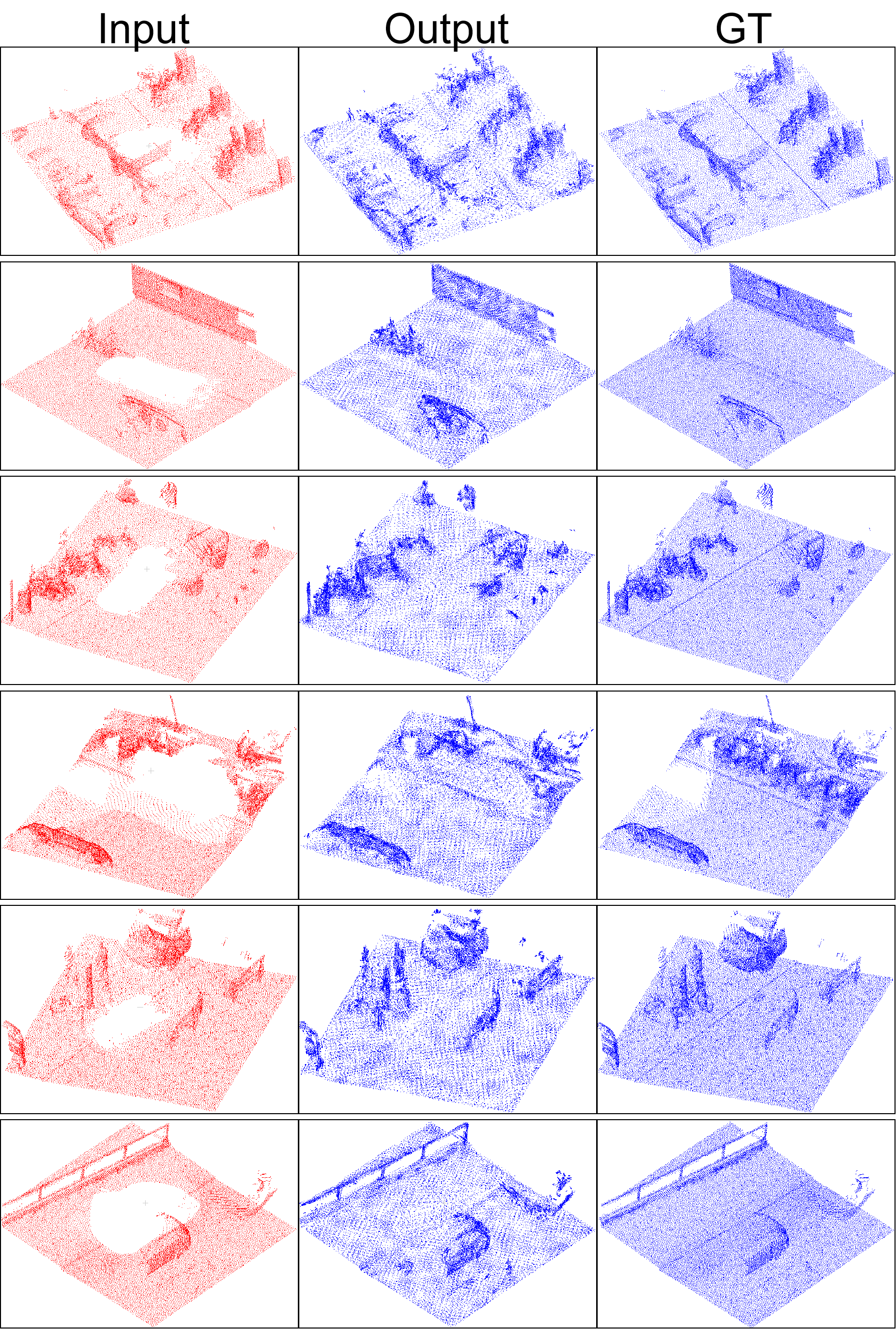}
\caption{Qualitative examples of completion results for scene with noise, including a significant amount of vegetation or scenes with parked bicycles. GT represents the ground truth of the 3D scene.}
\label{fig:Robustness of SGC-Net}
\end{figure*}

SGC-Net, as a point cloud completion network to be used in real-world scenarios, is expected to be robust and resistant to noise. 
Its performance should remain stable even in the presence of point cloud data that hasn't been well processed or with lower density. 
To demonstrate the robustness and noise resistance of the proposed SGC-Net, we present several scenes from the dataset that had a significant amount of noise.
These scenes include a significant amount of vegetation or scenes with parked bicycles, but they also have clear road boundaries. They were also included during the dataset generation.
As shown in Figure \ref{fig:Robustness of SGC-Net}, it is evident that even in the presence of interfering objects in real-world scenarios, the point cloud scenes can still be reliably completed.

\subsection{Ablation Study of SGC-Net}\label{sec:ablation}
The design of SGC-Net includes several important components. 
In order to investigate the effectiveness of different components in the network, we conducted an ablation study. 

It is worth noting that both the folding layer and the MLPs are used to densify the coarse point cloud, and at least one of these two modules must be used to generate a dense point cloud.
We designed several deep networks by including or excluding individual components:
(1) Model with z-axis enhancement and MLPs (GRNet gridding 80) without folding layers.
Z-axis enhancement refers to the practice of scaling values by a factor of 3 along the height or z-axis direction. 
This aims to make the height differences more pronounced, ensuring that the model can effectively capture even minor local variations in height.
(2) Model with z-axis enhancement and two folding layers without MLPs. 
(3) Model with MLPs and two folding layers without z-axis enhancement. 
(4) Model with z-axis enhancement, MLPs, and one folding layer. 
(5) Model with all components. 
We used the same data for training and tested these models with an output point number of 27,648. 
CD and F-Score are used as the evaluation metrics and the quantitative results of these metric tables are shown in Table \ref{table:Performance Comparison with Different Components}. 
From the table, we can see that the performance of the complete pipeline including all components is the best in all metrics.

\begin{figure*}[h]
\centering
\includegraphics[width=.8\textwidth]{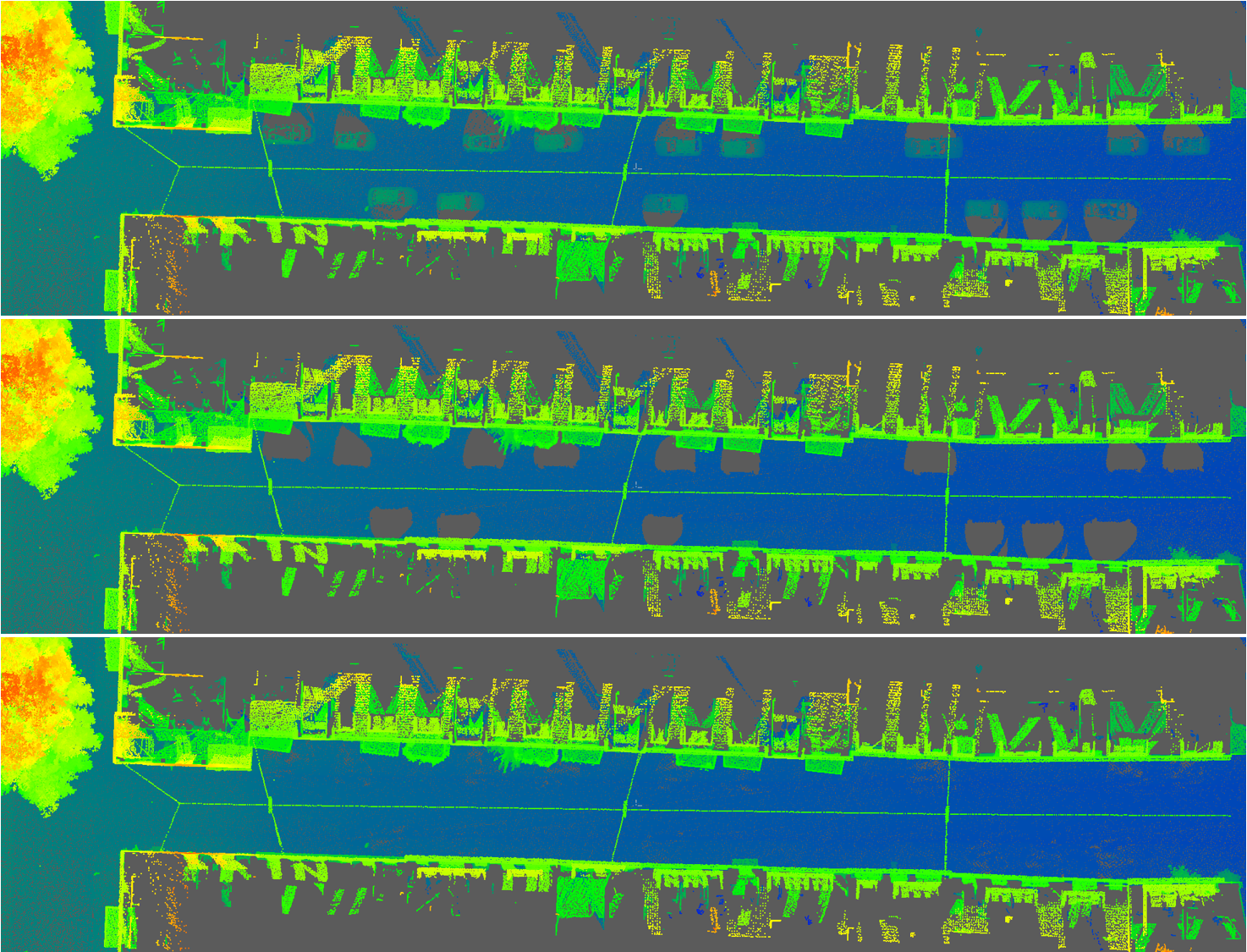}
\caption{Large-scale scene completion in the real world. The top figure is the original scene with vehicles parked on the roadside. The middle figure is the scene after removing the points of the vehicles, the gaps in the scene caused by the vehicle occlusion can be seen easily. The bottom figure is the result of the scene after the gap is filled by SGC-Net.}
\label{fig:Example of Large-Scale Scene Completion}
\end{figure*}

\begin{table*}[h!]
    \centering
    \caption{Ablation study on the performance of different components in the proposed method. The values in the table are the mean values of CD and F-Score calculated for each point in all point cloud scenes in the test set.}
    \label{table:Performance Comparison with Different Components}
    \begin{tabular*}{\tblwidth}{@{}CCCCCCC@{}}
    \toprule
        &\textbf{Z-Enhancement} & \textbf{MLPs} & \textbf{Folding 1} & \textbf{Folding 2} & \textbf{CD $\downarrow$} &  \textbf{F-Score $\uparrow$} \\ 
         & & & & & \textbf{($\times 10^{-4}$)} &  \\ \midrule
        (1)& $\checkmark$ & $\checkmark$ &  &  & 1.878 & 0.751 \\ 
        (2)& $\checkmark$ &  & $\checkmark$ & $\checkmark$ & 2.030 & 0.703  \\
        (3)& & $\checkmark$ & $\checkmark$ & $\checkmark$ & 2.856 & 0.501 \\
        (4)& $\checkmark$ & $\checkmark$ & $\checkmark$ &  & 1.797 & 0.753 \\
        (5)& $\checkmark$ & $\checkmark$ & $\checkmark$ & $\checkmark$ &\textbf{1.604} & \textbf{0.768} \\
    \bottomrule
    \end{tabular*}
\end{table*}

From the comparison of model (1), model (4) and model (5) in Table \ref{table:Performance Comparison with Different Components}, it can be seen that using the folding layer can improve the performance of the model, and adding one and two layers of folding layers can increase CD by 10.7\% and 14.6\%, making the F-Score increase by 1.5\% and 1.7\%. 
In particular, the CD error has been significantly reduced, which proves that the folding layer can better learn local features \citep{yang2018foldingnet}, making the generated results more refined and smooth locally. 
Comparing model (2) and model (5), we found that using the MLPs module has greatly improved the model, improving CD and F-Score by 21.0\% and 6.5\%, respectively. 
We conjecture that the features of each coarse cube can be downscaled by MLPs to retain their locally important features. The features output by MLPs should have distinct classification properties. 
The improvement in CD is particularly significant because the more useful features actually improve the fine detail of the completed point cloud. 

Compared with model (3), model (5) improves CD and F-Score by 43.8\% and 26.7\%, respectively, which confirms that enhancing the shape features of the Z axis is crucial for generating more accurate point clouds. 
This is because, in urban environments, the z-axis variations in road structures, especially road boundaries, are relatively minor compared to differences in xy dimensions.

For the completion of vertical walls and facades, gaps in the dataset are not as prominent, as they are typically much smaller in size compared to the gaps appearing on the ground.
Enlarging the scales in the z-axis direction can significantly enhance the network's performance in completing vertical shapes. 

\subsection{Point completion in the real-world scenarios}

\begin{figure*}[h!]
\centering
\includegraphics[width=.8\textwidth]{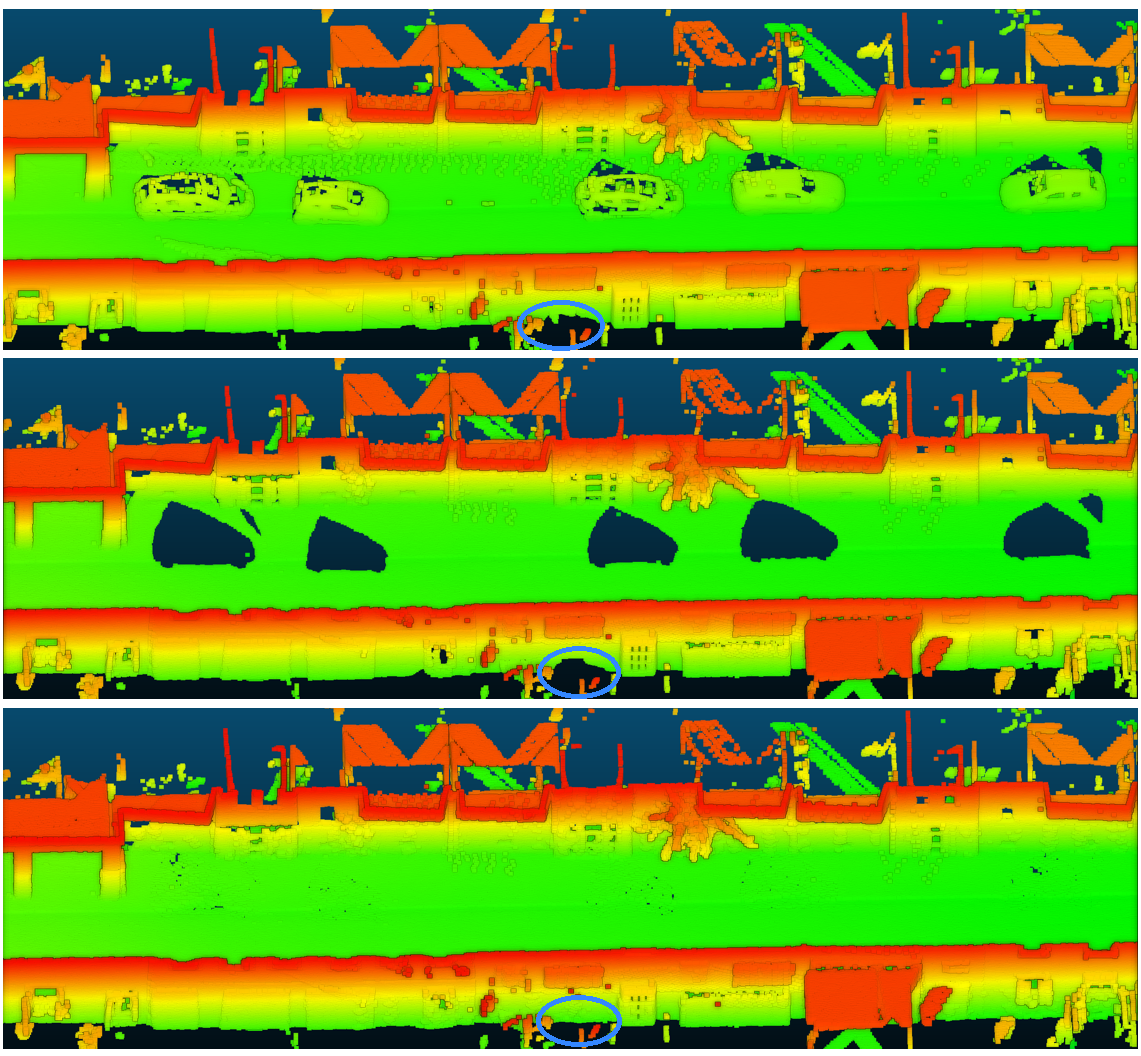}
\caption{Large-scale scene completion from an oblique perspective. The top figure is the original scene with vehicles parked on the roadside. The middle figure is the scene after removing the points of the vehicles. The bottom figure is the result of the scene after the gap is filled by SGC-Net.}
\label{fig:Example of Large-Scale Scene Completion from another Perspective}
\end{figure*}
Within our study area, we have chosen a road for validation based on real-world gap data. 
We used the SGC-Net trained in the previous step to fill the gaps caused by real vehicle occlusions. Thus we aim to verify the performance of our model in real-world scenarios.
We initiated the process by manually identifying the vehicle's location on the road and delineating a rectangular area measuring $8m \times 8m \times 2.6m$ around it. 
Subsequently, we manually removed the point cloud data belonging to the vehicles, as our focus was on the completion task, independent of detector performance. 
We assume that the points of the vehicles can be completely localized and removed before our model is used. 

Subsequently, we employed our SGC-Net to perform scene completion. Figure \ref{fig:Example of Large-Scale Scene Completion} present a top view of this road in its initial state (top), post-vehicle detection and removal (middle), and after gap completion (bottom).
Figure \ref{fig:Example of Large-Scale Scene Completion from another Perspective} provides an oblique view of the same process. It illustrates how our method effectively fills gaps, encompassing both ground and building facades, as highlighted within the blue circle.

\begin{figure*}[h!]
\centering
\includegraphics[width=0.75\textwidth]{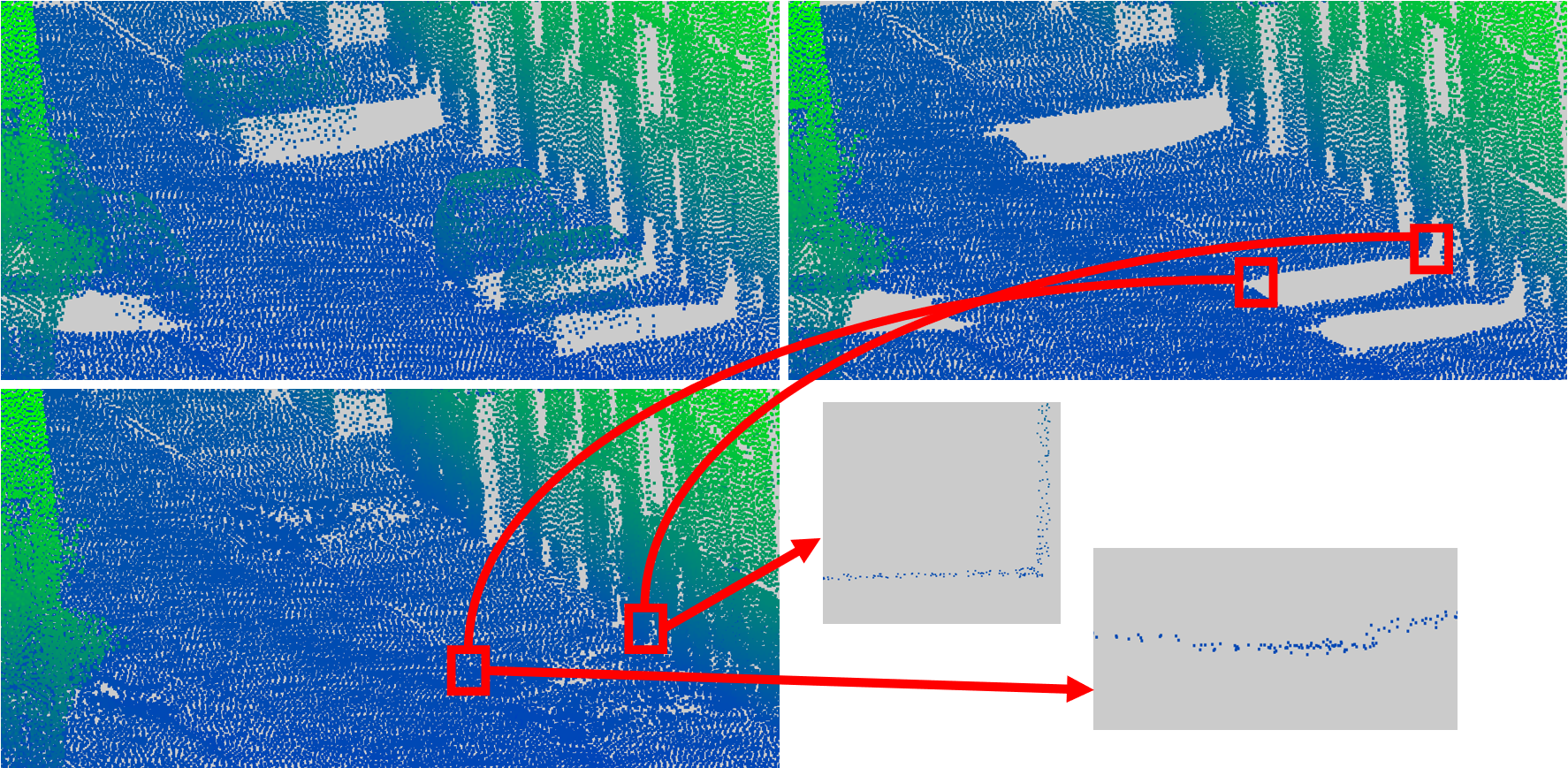}
\caption{Local details of the completed scene in real-world scenarios. The road boundaries and facades perpendicular to the ground are reasonably reconstructed.}
\label{fig:Local Details of the completed Scene}
\end{figure*}
\begin{figure}[h!]
\centering
\includegraphics[width=0.45\textwidth]{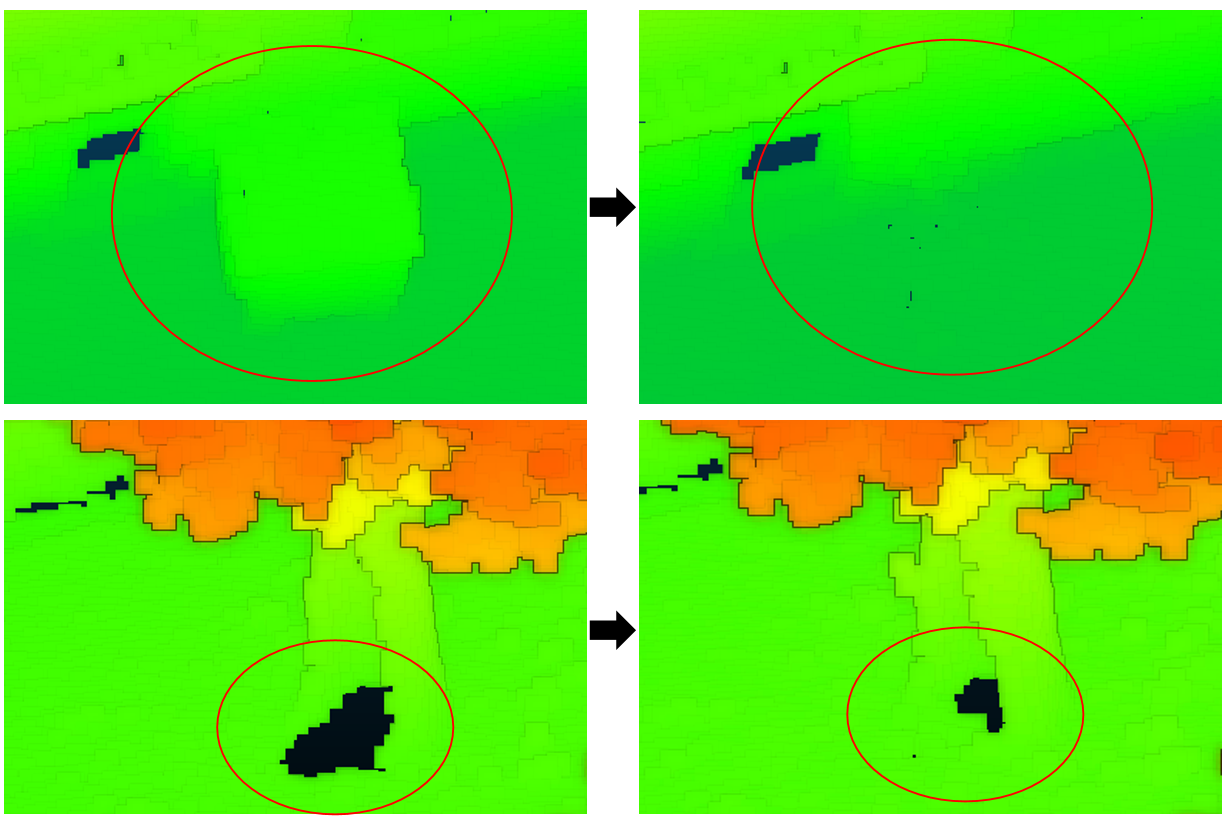}
\caption{Examples illustrating gap completion with SGC-Net. The top images demonstrate the filling of a gap previously obscured by a garbage can. Prior to SGC-Net completion, the garbage can points were manually removed. The bottom images depict the completion of a gap caused by a tree.}
\label{fig:Filling in Gaps caused by other Objects}
\end{figure}

In Figure \ref{fig:Local Details of the completed Scene}, we examine local details, revealing that SGC-Net fills gaps not only on the road surface but also on vertical facades. 
The completed point cloud has clear boundaries and geometric shapes, harmonizing with the surrounding scenes.
This can be attributed to two primary factors: the coarse point cloud generator can accurately reproduce the overall boundary shape of the scene, and then the local geometry can be well learned by folding.

Not only the gaps caused by vehicle occlusion can be filled by SGC-Net, but when we tested on real-world data, we found that the gaps caused by other objects can also be filled, such as garbage cans and trees. 
The reason for this would be due to the fact that the shapes of the gaps caused by different objects are similar. 
Figure \ref{fig:Filling in Gaps caused by other Objects} shows two examples where SGC-Net filled the gaps caused by garbage cans and trees, respectively.

\subsection{Discussion}

Our model performs well in the gap completion task which has also been verified using a real-world scenario.
Compared to existing studies that require deploying different workflows for individual scenarios, such as curbs in \cite{liu2022detection} or walls in \cite{gonzalez2024santiago}, the learning-based model proposed in this work showcases a more flexible approach, capable of handling different scenarios within a single model.

Nevertheless, there are still several limitations that could be addressed in further work. 
In this work, several thresholds need to be selected, such as those required in the data preparation step.
Although they are thresholds that only affect the selection of training data and are relatively insensitive to the trained model, there remains potential for further optimization.
Furthermore, our current work primarily concentrates on the challenge of gap completion. However, further advancements in automatically identifying gaps in point clouds would contribute to achieving full process automation.

Moreover, in this work, we only considered occlusions caused by common-sized vehicles, while other traffic participants (e.g., pedestrians, cyclists), trees, traffic furniture, trams, and very large vehicles were not yet taken into consideration. 
In addition, our current focus is mainly on mobile mapping data, where its adaptability to different scanning devices and their various mounting methods would also need further investigation.





\section{Conclusion}

In this work, we introduce SGC-Net, a deep neural network specifically designed for filling gaps in urban point cloud scenes that arise due to vehicle occlusion. 
The motivation behind this study stems from the inevitable occlusions caused by the numerous vehicles in urban environments during 3D data collection through mobile mapping systems.
This poses significant challenges for subsequent tasks, such as terrain or building modeling.

We leveraged ray-casting to automatically generate an extensive training dataset, effectively tackling the problem of lacking training data for point completion purposes.
Furthermore, our novel neural network architecture, SGC-Net, surpasses existing methods in terms of gap-filling and point cloud scene reconstruction.
SGC-Net generates detailed point cloud scenes and completes real-world point cloud scenes with clear geometric boundaries. 
The model's effectiveness and design rationale are demonstrated through comparative and ablation experiments. 

As for future work, we have identified four aspects from the limitations we discussed: 
(1)~our method could be extended to complete gaps in building facades obscured by various objects, such as trees, billboards, streetlights, etc., enhancing the completion of 3D point clouds; 
(2)~it is worth further exploring whether this workflow can be directly transferred to other scenarios to generate training data, such as using different scanning devices (e.g., Velodyne, Blickfeld, etc.) and their various mounting methods (e.g., horizontally on the roof, vertically on the rear);
(3)~due to the constraints of GPU graphics memory, only small to medium-sized vehicles were considered for generating training examples. Nonetheless, addressing occlusion from larger vehicles would require larger input scenes and more capable hardware, offering a more effective solution; and 
(4)~if semantic segmentation information is available, such as road markings, road boundary, grass, and building facades in the point cloud scene, it has the potential to further enhance gap completion and provide sharper geometric boundaries.

\bibliographystyle{cas-model2-names}

\bibliography{cas-refs}

\end{document}